\documentclass{article}

\usepackage{microtype}
\usepackage{graphicx}
\usepackage{subcaption}
\usepackage{booktabs} %

\usepackage{hyperref}

\usepackage[preprint]{icml2026}

\usepackage{amsmath}
\usepackage{amssymb}
\usepackage{mathtools}
\usepackage{amsthm}

\usepackage[capitalize,noabbrev]{cleveref}

\theoremstyle{plain}

\theoremstyle{definition}

\theoremstyle{remark}

\usepackage[textsize=tiny]{todonotes}

\usepackage{csquotes}
\usepackage{silence}
\usepackage{multirow}
\usepackage[table]{xcolor}

\icmltitlerunning{LKV: End-to-End Learning of Head-wise Budgets and Token Selection for LLM KV Cache Eviction}

\begin{document}

\twocolumn[
  \icmltitle{LKV: End-to-End Learning of Head-wise Budgets and Token Selection for LLM KV Cache Eviction}

  \begin{icmlauthorlist}
    \icmlauthor{Enshuai Zhou}{ustc,ict}
    \icmlauthor{Yifan Hao}{ict}
    \icmlauthor{Chao Wang}{ustc}
    \icmlauthor{Rui Zhang}{ict}
    \icmlauthor{Di Huang}{ict}
    \icmlauthor{Jiaming Guo}{ict}
    \icmlauthor{Xing Hu}{ict}
    \icmlauthor{Zidong Du}{ict}
    \icmlauthor{Qi Guo}{ict}
    \icmlauthor{Yunji Chen}{ict,ucas}
  \end{icmlauthorlist}

  \icmlaffiliation{ustc}{University of Science and Technology of China, Hefei, China}
  \icmlaffiliation{ict}{State Key Lab of Processors, Institute of Computing Technology, CAS, Beijing, China}
  \icmlaffiliation{ucas}{University of Chinese Academy of Sciences, Beijing, China}

  \icmlcorrespondingauthor{Enshuai Zhou}{enszhou@mail.ustc.edu.cn}
  \icmlcorrespondingauthor{Yunji Chen}{cyj@ict.ac.cn}

  \icmlkeywords{Machine Learning}

  \vskip 0.3in
]

\printAffiliationsAndNotice{}  %

\newcommand{\methodname}{\textsc{LKV}}
\newcommand{\methodnameH}{\textsc{LKV-H}}
\newcommand{\methodnameT}{\textsc{LKV-T}}

\newcommand{\sysname}{\textsc{DeepFlow}}

\newcommand{\R}{\mathbb{R}}
\newcommand{\loss}{\mathcal{L}}
\newcommand{\vct}[1]{\mathbf{#1}} %

\begin{abstract}
    Long-context inference in Large Language Models (LLMs) is bottlenecked by the linear growth of Key-Value (KV) cache memory. Existing KV cache compression paradigms are fundamentally limited by heuristics: heuristic budgeting relies on statistical priors rather than task objectives, causing resource misallocation, while heuristic selection relies on coupled query-key interactions or static inductive biases (e.g., attention sinks). To address this limitation, we introduce LKV (Learned KV Eviction), which formulates KV compression as an end-to-end differentiable optimization problem. LKV integrates LKV-H to learn task-optimized global budgets, and LKV-T to derive intrinsic KV importance without materializing attention matrices. This design bypasses heuristic proxies, strictly aligning compression with task objectives. Extensive evaluations demonstrate that LKV achieves state-of-the-art performance on both LongBench and RULER benchmarks at high compression rates. In particular, on LongBench, LKV achieves near-lossless performance with only 15\% KV cache retention. Crucially, our analysis identifies learned budgeting as the dominant driver of fidelity, demonstrating that data-driven allocation is essential to overcome the limitations of hand-crafted heuristics.
\end{abstract}

\section{Introduction}
\label{sec:intro}

Processing long contexts is essential for Large Language Models (LLMs) applications
like repository-level code analysis~\citep{RepoCoderRepositoryLevelCode-2023} and multi-document question answering~\citep{LostMiddleHow-2024}.
However, processing these extensive inputs imposes a prohibitive \textit{memory wall}~\citep{AIMemoryWall-2024a,FlashAttentionFastMemoryEfficient-2022}.
As the model encodes a long sequence, the accumulation of Key-Value (KV) states triggers an immediate bottleneck~\citep{EfficientlyScalingTransformer-2023,EfficientMemoryManagement-2023}, which severely limits the maximum handleable context length and drastically reduces inference throughput on memory-constrained hardware~\citep{FlexGenHighthroughputGenerative-2023a}.

To overcome this barrier, KV Cache Eviction has emerged as a vital technique~\citep{ScissorhandsExploitingPersistence-2023,H2OHeavyHitterOracle-2023}.
The core principle involves dynamically identifying and discarding non-essential tokens while retaining critical information, thereby mitigating memory growth as sequence length increases.
By aggressively compressing the KV cache, eviction significantly reduces memory footprints for applications such as high-throughput serving and on-device deployment, enabling models to sustain high-performance generation over extended contexts within finite memory budgets~\citep{ModelTellsYou-2023,SnapKVLLMKnows-2024}.

Current approaches, however, remain constrained by interconnected limitations across budget allocation and token selection.
First, early methods like H2O~\citep{H2OHeavyHitterOracle-2023} and SnapKV~\citep{SnapKVLLMKnows-2024} largely adopt uniform budgets, ignoring the significant heterogeneity of attention heads where critical information is sparse and non-uniformly distributed.
Second, while advanced strategies attempt to address this, they are trapped by rigid priors or costly proxies.
PyramidKV~\citep{PyramidKVDynamicKV-2025} enforces a static layer-wise decay that fails to accommodate intermediate retrieval heads.
Meanwhile, adaptive methods like Ada-KV~\citep{AdaKVOptimizingKV-2025} and D2O~\citep{D2ODynamicDiscriminative-2025} rely on calculating attention scores to allocate budgets or select tokens.
This creates an inference-time circular dependency: identifying important components requires performing the heavy query-key interactions ($O(t^2)$ complexity) that we aim to avoid, preventing true query-agnostic acceleration~\citep{KeyformerKVCache-2024}.
Third, and most critically, optimizing these \textit{proxy metrics} (attention weights) rather than the task objective often necessitates manual safeguards to ensure stability.
For budget allocation, methods frequently enforce generic constraints (e.g., minimum budget thresholds)~\citep{AdaKVOptimizingKV-2025,NotAllHeads-2025} to prevent over-pruning specific heads.
For token selection, they rely on fixed \enquote{recent windows} or \enquote{attention sinks}~\citep{EfficientStreamingLanguage-2023,TamingFragilityKV-2025} to maintain fluency.
While these generic priors act as effective stabilizers, they impose a prescriptive structure that constrains the model from autonomously discovering complex sparsity patterns that deviate from hand-crafted rules.

We posit that optimal KV compression should be an end-to-end learned capability rather than a collection of heuristic rules.
Instead of relying on manual rules, the task objective itself should directly guide the optimization of both macro-level budget allocation (e.g., distinguishing critical retrieval heads from streaming heads) and micro-level token utility (e.g., balancing distant anchors against recent local context).
We introduce \textbf{\methodname{}} (\textbf{L}earned \textbf{KV} \textbf{E}viction),
which, to the best of our knowledge, is the first framework that reformulates KV cache compression as a unified, end-to-end differentiable optimization problem.

\begin{figure}[t]
    \centering
    \includegraphics[width=1.0\linewidth]{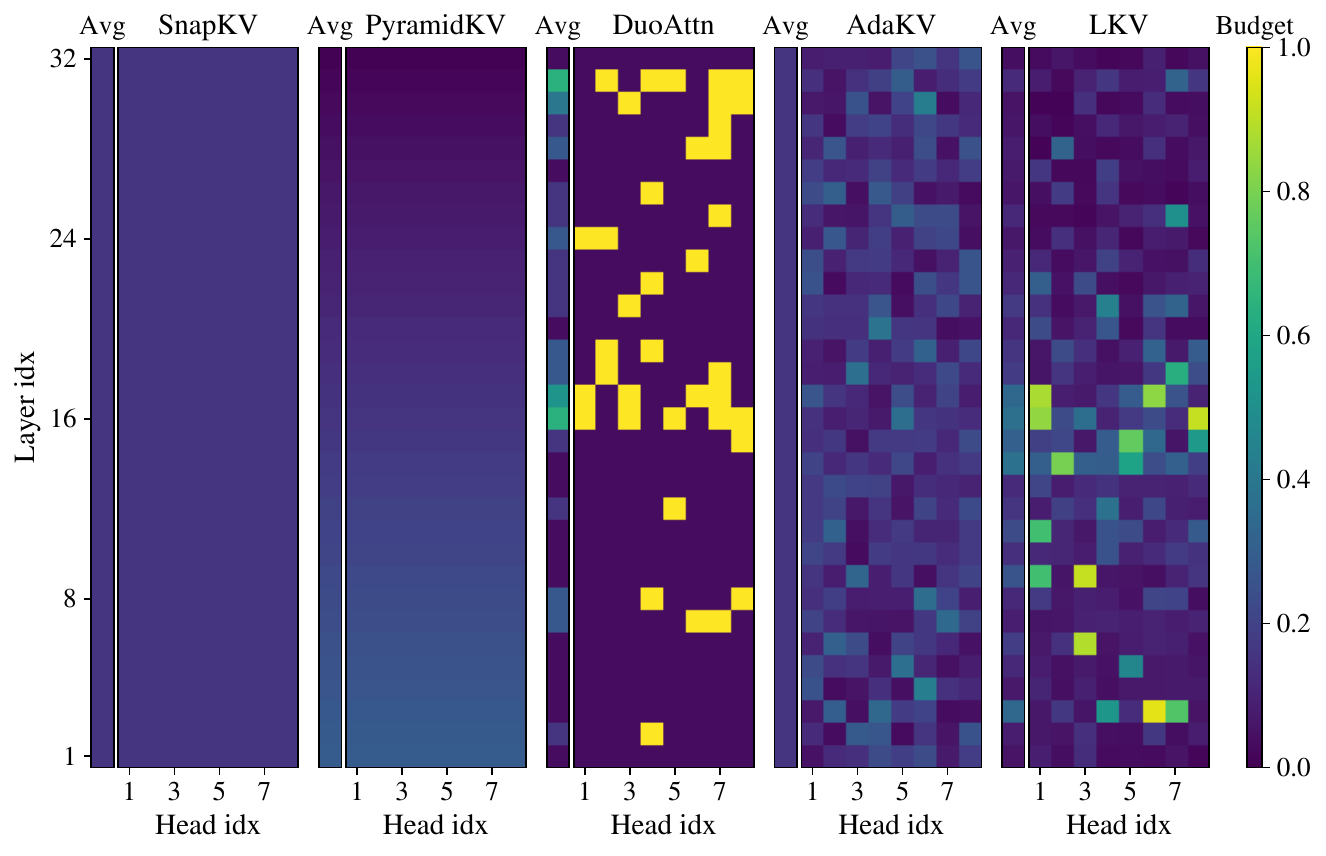}
    \caption{KV budget allocation policies (15\% retention).
        \textbf{(a)} SnapKV: Uniform.
        \textbf{(b)} PyramidKV: Layer-wise decay.
        \textbf{(c)} DuoAttention: Rigid binary classification (Retrieval vs. Streaming).
        \textbf{(d)} Ada-SnapKV: Adaptive within layers but with uniform layer priors.
        \textbf{(e)} \methodname{} (Ours): Learned global, fine-grained policy optimizing task objectives without rigid priors.}
    \label{fig:teaser}
\end{figure}

As visualized in Figure~\ref{fig:teaser}, \methodname{} reshapes resource allocation for all KV heads.
Unlike uniform (Figure~\ref{fig:teaser}a), rigid structural priors (Figure~\ref{fig:teaser}b-c) enforcing layer-wise decay or strict binary dichotomies, or heuristic adaptation (Figure~\ref{fig:teaser}d) relying on noisy proxies, \methodname{} learns a globally optimized, constraint-free policy that directs budgets to the most critical heads purely on demand.
\textbf{\methodnameH{}} (Global Learned Budgeting) enables global competition by flattening the head structure, allowing every head to compete for the total budget via learnable embeddings.
This breaks the \enquote{pyramidal} trap: as shown in Figure~\ref{fig:teaser}d, \methodname{} freely assigns high budgets to deeper layers if the task demands, whereas Ada-KV remains confined by static decay.
The allocation is learned entirely from data, without relying on heuristics or safeguard hyperparameters.
Simultaneously, \textbf{\methodnameT{}} (Matrix-Free Selection) predicts token utility directly from KV states via a lightweight network.
This design enables matrix-free selection with negligible linear $O(t)$ overhead, avoiding costly $O(t^2)$ attention computation.
Notably, \methodname{} operates layer-by-layer during the prefill phase, effectively evicting tokens to lower peak memory consumption.

We train \methodname{} end-to-end via self-distillation~\citep{BeYourOwn-2019}.
Extensive experiments on LongBench~\citep{LongBenchBilingualMultitask-2024} and RULER~\citep{RULERWhatsReal-2024} demonstrate that \methodname{} significantly outperforms state-of-the-art baselines.
Our analysis further reveals that the \textit{learned global budget} contributes more to performance gains than the selection policy itself, demonstrating the importance of precise, head-level resource allocation. Our contributions are summarized as follows:
\begin{itemize}
    \item We propose \methodname{}, the first framework to transform KV eviction into an end-to-end differentiable learning problem, eliminating the need for hand-crafted heuristics, minimum budgets, or forced recent windows.
    \item We introduce \methodnameH{}, a global budgeting mechanism that uncovers intrinsic head importance, and \methodnameT{}, a matrix-free selector for efficient inference.
    \item We achieve SOTA performance on long-context benchmarks and provide new insights: learned budgeting is the dominant factor in compression quality.
\end{itemize}

\section{Related Work}

\subsection{Eviction-based KV Cache Compression}

Eviction-based KV cache compression primarily focuses on two aspects: strategies for budget allocation and mechanisms for token selection.

\paragraph{Budget Allocation Strategies.}
The distribution of the KV cache budget significantly impacts model fidelity.
Early methods typically employ uniform allocation across all layers and heads, such as H2O~\citep{H2OHeavyHitterOracle-2023} and SnapKV~\citep{SnapKVLLMKnows-2024}.
Recognizing that layers contribute unequally to information processing~\citep{RethinkingValueTransformer-2020,LayerLayerUncovering-2025}, recent works have introduced non-uniform strategies.
PyramidKV~\citep{PyramidKVDynamicKV-2025} and PyramidInfer~\citep{PyramidInferPyramidKV-2024} employ a pyramidal strategy, prioritizing shallower layers with higher cache budgets. Similarly, D2O~\citep{D2ODynamicDiscriminative-2025} adjusts budgets dynamically according to layer-wise attention density. Transitioning to finer granularity, Ada-KV~\citep{AdaKVOptimizingKV-2025} tunes budget distribution across different heads within each layer. Further extending this flexibility, CAKE~\citep{CAKECascadingAdaptive-2025} treats cache allocation as a cascading cake-slicing task based on spatial-temporal attention patterns.
While these methods rely on statistical or rule-based heuristics, our LKV-H module treats macro-budgeting as an end-to-end differentiable optimization problem, enabling a global competition for resources that bypasses rigid layer-wise priors.

\paragraph{Token Selection Mechanisms.}
Once budgets are established, the model must decide which tokens to retain.
Static strategies like StreamingLLM~\citep{EfficientStreamingLanguage-2023} prioritize initial "attention sinks" and recent tokens~\citep{LLMsKnowWhat-2025,LocretEnhancingEviction-2025} to stabilize decoding.
Heuristic dynamic methods often assess token importance through coupled query-key interactions; for instance, H2O~\citep{H2OHeavyHitterOracle-2023} tracks cumulative attention scores, while SnapKV~\citep{SnapKVLLMKnows-2024} leverages a local observation window to identify key clusters.
Recent advancements have introduced value-aware metrics, such as CriticalKV~\citep{IdentifyCriticalKV-2025}, which considers the norm of projected value states to bound output perturbation.
Furthermore, learning-based approaches like DuoAttention~\citep{DuoAttentionEfficientLongContext-2024} and PruLong~\citep{CacheMeIf-2025} attempt to identify retrieval heads through distillation or optimization.
However, most existing mechanisms require materializing the attention matrix to evaluate importance.
In contrast, LKV-T employs a matrix-free selection process that predicts token utility from intrinsic latent features, eliminating the need for expensive query-dependent attention calculations.

\subsection{Other LLM Efficiency Paradigms}
Our work is orthogonal to other paradigms, including sparse attention (e.g., Quest~\citep{QuestQueryAwareSparsity-2024}, MInference~\citep{MInference10Accelerating-2024}), prompt compression (e.g., LLMLingua-2~\citep{LLMLingua2DataDistillation-2024}, 500xCompressor~\citep{500xCompressorGeneralizedPrompt-2024}), and KV quantization (e.g., KIVI~\citep{KIVITuningFreeAsymmetric-2024}, AQUA-KV~\citep{CacheMeIf-2025a}). Unlike these methods which focus on selective loading, input shrinkage, or precision reduction, LKV reduces cardinality and can be integrated with them for complementary gains.

\section{Methodology}
\label{sec:method}

\begin{figure*}[t]
    \centering
    \includegraphics[width=1.0\linewidth]{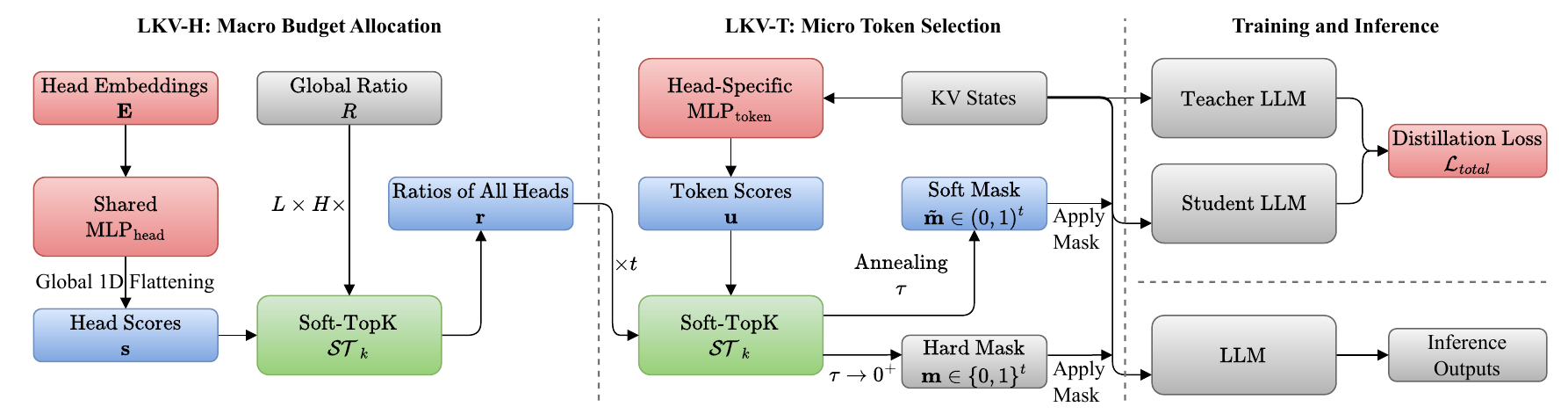}
    \caption{Overview of \methodname{}. \textbf{Left (LKV-H):} Learns global budget ratios $\mathbf{r}$ from head embeddings. \textbf{Middle (LKV-T):} Performs differentiable, query-agnostic token selection via Soft-TopK. \textbf{Right:} End-to-end optimization via self-distillation against a frozen teacher.}
    \label{fig:framework}
\end{figure*}

\subsection{Preliminaries and Problem Formulation}
\label{subsec:preliminaries}

Consider a Decoder-only Large Language Model (LLM) with $L$ layers. We focus our formulation on the \textbf{Key-Value (KV) heads}, as they constitute the memory bottleneck. Let $H$ denote the number of KV heads per layer. This generalized setting naturally accommodates both multi-head attention (MHA) and grouped-query attention (GQA), where multiple query heads may share a single KV head~\citep{AttentionAllYou-2017,GQATrainingGeneralized-2023}. Given an input sequence of tokens $x_1, \dots, x_t$, the model generates the next token $x_{t+1}$ in an autoregressive manner.

\paragraph{The KV Cache Bottleneck.}
During autoregressive generation at step $t$, the model maintains a KV cache $\mathcal{C}_t = \{(\mathbf{K}^{(l,h)}_{\le t}, \mathbf{V}^{(l,h)}_{\le t})\}_{l=1, h=1}^{L, H}$ to avoid redundant computations. Here, $\mathbf{K}, \mathbf{V} \in \mathbb{R}^{t \times d}$ represent the stacked key and value states for the $h$-th KV head in layer $l$. As the sequence length $t$ increases, the memory footprint of $\mathcal{C}_t$ grows linearly, imposing a significant \textit{memory wall} on system throughput. The standard attention output for a query $\mathbf{q}_t$ attending to this cache is computed as:
\begin{equation}
    \label{eq:attention}
    \mathbf{o}_{t} = \text{Softmax}\left(\frac{\mathbf{q}_t (\mathbf{K}_{\le t})^\top}{\sqrt{d}}\right) \mathbf{V}_{\le t}.
\end{equation}

\paragraph{Problem Formulation.}
Our objective is to compress $\mathcal{C}_t$ by learning a binary retention mask $\mathbf{M} \in \{0, 1\}^{L \times H \times t}$ for all KV cache items. Crucially, this mask operates at the KV-head level, ensuring that any eviction decision is consistently applied across all query heads sharing that KV cache. We define a target global retention ratio $R \in (0, 1)$, which determines the total KV budget at step $t$ as $B_{\text{total}} = \lfloor R \cdot L \cdot H \cdot t \rfloor$. We formulate \methodname{} as an end-to-end optimization problem to minimize the discrepancy between the full-cache and compressed models:
\begin{equation}
    \label{eq:objective}
    \begin{aligned}
         & \min_{\phi} \quad \mathbb{E}_{x \sim \mathcal{D}} \left[ \mathcal{L} \left( \text{LLM}(\mathcal{C}_{\text{full}}), \text{LLM}(\mathbf{M} \odot \mathcal{C}_{\text{full}}; \phi) \right) \right] \\
         & \text{s.t.} \quad \|\mathbf{M}\|_0 = \sum_{l=1}^{L} \sum_{h=1}^{H} \|\mathbf{m}^{(l,h)}\|_0 \le B_{\text{total}}.
    \end{aligned}
\end{equation}
where $\phi$ represents the learnable parameters of \methodname{}, and $\mathcal{L}$ denotes the distillation loss function, which will be detailed in Section~\ref{subsec:training}.
The overall framework of \methodname{} is illustrated in Figure~\ref{fig:framework}.
We now describe the core components of our method in detail.

\subsection{The Differentiable Soft-TopK Operator}
\label{subsec:soft_topk}

End-to-end learning requires a differentiable approximation of the discrete Top-$k$ selection. Standard heuristic methods often rely on non-differentiable sorting or local statistics, which block the gradient flow required for optimizing the budget allocation and selection policy. To bridge this gap, we adopt the \textbf{Soft-TopK Operator}, denoted as $\mathcal{ST}_k(\cdot)$, which is based on the smooth approximation framework explored by~\citet{PostSoftmaxSearchingSmooth-2024}. Unlike simple temperature-scaled approximations, this formulation guarantees rigorous constraint satisfaction and numerical stability.

\paragraph{Mathematical Definition.}
Let $\vct{x} \in \mathbb{R}^n$ be the input score vector and $k \in (0, n)$ be the target parameter. We define the element-wise Soft-TopK operator with an input-dependent threshold $\lambda(\vct{x})$ as:
\[
    \mathcal{ST}_k(\vct{x})_i \triangleq f(x_i - \lambda(\vct{x})) \quad \text{s.t.} \quad \sum_{i=1}^n f(x_i - \lambda(\vct{x})) = k.
\]
We employ the CDF of the Laplace distribution as $f$ for its symmetric gradients and analytical tractability: $f(z) = \frac{1}{2}(1 + \text{sgn}(z)(1 - e^{-|z|}))$. Consequently, the core challenge lies in analytically determining $\lambda(\vct{x})$ such that the summation constraint is strictly satisfied.

\paragraph{Closed-Form Solution for $\lambda$.}
To strictly enforce the budget constraint, we derive an analytical solution for the threshold $\lambda(\mathbf{x})$.
Assuming sorted inputs, the constraint $\sum f(x_i - \lambda) = k$ yields a quadratic equation in terms of $e^{\lambda}$. The valid root provides the explicit closed-form:
\begin{equation}
    \lambda(\mathbf{x}) = \log \left( \frac{\sqrt{(k-m)^2 + S_1 S_2} - (k-m)}{S_1} \right),
\end{equation}
where the partial sums are defined as $S_1 = \sum_{i=1}^m e^{-x_i}$ and $S_2 = \sum_{i=m+1}^n e^{x_i}$.
This analytical formulation guarantees exact budget satisfaction and maintains end-to-end differentiability with respect to both input scores $\mathbf{x}$ and the budget parameter $k$.
Detailed derivation, numerical stability analysis, and gradient proofs are provided in Appendix~\ref{app:soft_topk}.

\paragraph{Theoretical Properties.}
This formulation constructs a mapping $\mathcal{ST}_k: \mathbb{R}^n \to (0, 1)^n$ satisfying three properties:
\begin{enumerate}
    \item Normalization: The output vector always sums to $k$, enforcing the budget constraint.
    \item Order Preservation: $x_i > x_j \implies \mathcal{ST}_k(\vct{x})_i \ge \mathcal{ST}_k(\vct{x})_j$, ensuring prioritization of important tokens.
    \item Translation Invariance: $\mathcal{ST}_k(\vct{x} + c) = \mathcal{ST}_k(\vct{x})$. This decouples selection from absolute magnitude, stabilizing training.
\end{enumerate}

\paragraph{Theoretical Unification.}
Our framework bridges established selection mechanisms through a generalized perspective. A temperature parameter $\tau > 0$ can be naturally introduced (via input scaling $\vct{x}/\tau$) to adjust the sharpness of the distribution. This reveals two critical limiting cases: the operator approximates to the standard Softmax function when $k=1$ (at $\tau=1$), and converges to the discrete Hard Top-$k$ selection in the limit as $\tau \to 0^+$.

\subsection{\methodnameH{}: Learning Global Budget Allocation}
\label{subsec:lkv_h}

A critical limitation of existing compression paradigms is their fragmented scope of budget allocation. Heuristic methods typically operate within restricted structural boundaries: methods like Ada-KV focus solely on \textit{intra-layer} redistribution, while approaches like PyramidKV enforce fixed \textit{inter-layer} decay patterns. Achieving a holistic allocation often requires combining multiple heuristics, which is suboptimal and complex.
In reality, the utility of attention heads is not confined by layer boundaries.

\methodnameH{} removes these constraints by formulating budget allocation as a \textbf{unified, global competition}. Instead of optimizing layer-wise or head-wise separately, we flatten the entire model's attention structure into a single dimension, allowing every head to compete for the global memory capacity directly.

\paragraph{Head Embeddings and Scoring.}
We assign a learnable embedding vector $\mathbf{e}^{(l,h)} \in \mathbb{R}^{d_e}$ to each KV head. A shared MLP maps these embeddings to scalar importance scores:
\begin{equation*}
    s^{(l,h)} = \text{MLP}_{\text{head}}(\mathbf{e}^{(l,h)}).
\end{equation*}
Unlike statistical proxies (e.g., attention concentration) used in prior works, these scores quantify the intrinsic contribution of each head to the training objective.

\paragraph{Global 1D Competition.}
Let $R \in (0, 1)$ be the target global retention ratio.
The aggregate retention ratio (sum of ratios among all heads) is $R_{\text{agg}} = R \cdot L \cdot H$.
Crucially, we bypass hierarchical rigidities by flattening the score tensor $\mathbf{s}=\{s^{(l,h)}\}^{L \times H}$ into a one-dimensional vector $\mathbf{s}_{\text{flat}} \in \mathbb{R}^{LH}$.
This formulation transforms allocation into a global competition, enabling \textit{cross-layer budget transfer}: a critical head in the deepest layer can effectively claim capacity from a redundant head in the first layer, forcing resources to align strictly with information density rather than structural depth.
We then apply the Soft-TopK operator (with $\tau=1$) to this global vector:
\begin{equation*}
    \label{eq:budget_allocation}
    \mathbf{r}_{\text{flat}} = \mathcal{ST}_{R_{\text{agg}}}(\mathbf{s}_{\text{flat}}), \quad \text{where} \;\; \sum_{i=1}^{L \cdot H} (\mathbf{r}_{\text{flat}})_i = R_{\text{agg}}.
\end{equation*}
By reshaping $\mathbf{r}_{\text{flat}}$ back to $\mathbf{r} \in \mathbb{R}^{L \times H}$, we obtain the specific retention ratio $r^{(l,h)} \in (0, 1)$ for each head. This \textit{flattened optimization} enables cross-layer resource transfer: the model can automatically discover complex patterns—such as allocating heavy budgets to middle-layer retrieval heads while sparsifying both initial and final layers—that rigid layer-wise heuristics can never approximate.

\paragraph{Zero-Overhead Budgeting at Inference.}
Since the learned ratios $\mathbf{r}$ depend solely on static embeddings, they are pre-computed once and frozen. At inference step $t$, the discrete token budget for head $(l,h)$ is simply $b^{(l,h)} = \lfloor r^{(l,h)} \cdot t \rfloor$, incurring zero runtime overhead. With the precise budget established, the remaining challenge is to select specific tokens, which we address next.

\subsection{\methodnameT{}: Matrix-Free Intrinsic Token Selection}
\label{subsec:lkv_t}

Having established the optimal capacity (budget) for each head via \methodnameH{}, the framework must now determine the specific content (tokens) to retain. While prior methods like H2O and SnapKV rely on query-key attention scores to assess token utility, they suffer from a "chicken-and-egg" problem: evaluating utility requires computing the heavy attention matrix that compression aims to avoid.

\methodnameT{} resolves this by introducing a \textbf{Matrix-Free} selection mechanism. We posit that a token's utility is intrinsic to its latent features and can be predicted independently of the current query.

\paragraph{Head-Specific Intrinsic Scoring.}
To capture the distinct sparsity patterns of different attention heads, we assign a unique, lightweight MLP to each head. For the $i$-th token in head $h$ of layer $l$, we concatenate its key $\mathbf{k}_i^{(l,h)}$ and value $\mathbf{v}_i^{(l,h)}$ to predict an intrinsic importance score $u_i$:
\begin{equation*}
    u_i^{(l,h)} = \text{MLP}_{\text{token}}^{(l,h)}([\mathbf{k}_i^{(l,h)}; \mathbf{v}_i^{(l,h)}]).
\end{equation*}
This scoring is \textit{query-agnostic}, meaning the importance is derived solely from the KV states themselves, avoiding dependency on the current query $\mathbf{q}_t$.

\paragraph{Hierarchical Application of Soft-TopK.}
Note that the Soft-TopK operator here differs from its usage in \methodnameH{} (Sec.~\ref{subsec:lkv_h}).
In \methodnameH{}, the operator is used to \textit{generate the retain ratios} $\mathbf{r}=\{r^{(l,h)}\}$ for all heads.
Subsequently in \methodnameT{}, we employ the operator to \textit{select tokens} based on the allocated budget.
To enable joint optimization, we feed the continuous budget $b^{(l,h)} = r^{(l,h)} \cdot t$ into the operator during training, allowing gradients to propagate from token selection to the budget allocation due to differentiability with respect to $b$.
We apply the discrete floor operation $b^{(l,h)} = \lfloor r^{(l,h)} \cdot t \rfloor$ only at inference time.

\paragraph{Gumbel-Based Relaxation and Annealing.}
A major challenge in learning binary masks is the discrepancy between the differentiable soft mask used during training and the discrete hard selection required for inference. To bridge this gap, we adopt a Gumbel-Noise relaxation combined with temperature annealing.

During \textbf{training}, to encourage exploration and differentiability, we inject Gumbel noise $\mathbf{g} \sim \text{Gumbel}(0, 1)$ into the scores and apply the Soft-TopK operator with a temperature parameter $\tau$:
\begin{equation*}
    \tilde{\mathbf{m}} = \mathcal{ST}_{b}(\mathbf{u} + \mathbf{g}; \tau), \quad \text{where} \;\;\sum_{i=1}^t \tilde{m}_i = b.
\end{equation*}
Here, $\tilde{\mathbf{m}} \in (0, 1)^t$ represents the continuous relaxation of the retention mask. The temperature $\tau$ controls the sharpness of the distribution. We employ a \textit{temperature annealing schedule}, gradually decaying $\tau$ from a high value (high randomness, soft selection) to a near-zero value (deterministic, hard selection) over the course of training. This strategy allows the model to explore the search space early on and converge to a discrete solution that closely mimics the inference behavior.

During \textbf{inference}, we remove the stochasticity and soft approximation. We perform a deterministic Hard Top-K selection on the raw scores $\mathbf{u}$ to retain exactly $b$ tokens:
\begin{equation*}
    \mathbf{m} = \text{TopK}(\mathbf{u}, b), \quad \mathbf{m} \in \{0, 1\}^t.
\end{equation*}
Thanks to the annealing process, the distributional shift between the training mask $\tilde{\mathbf{m}}$ (at late stages) and the inference mask $\mathbf{m}$ is minimized.

\subsection{Training Objective}
\label{subsec:training}

\paragraph{Distillation Loss.}
We train \methodname{} via self-distillation while keeping the LLM backbone entirely \textbf{frozen}.
The Student ($S$, compressed) learns to mimic the Teacher ($T$, full-cache) by minimizing a joint objective that aligns both output distributions and intermediate representations:
\begin{equation*}
    \mathcal{L}_{\text{total}} = D_{\text{KL}}(P_T \parallel P_S) + \frac{\beta}{L} \sum_{l=1}^{L} \| \mathbf{H}^{(l)}_{\text{T}} - \mathbf{H}^{(l)}_{\text{S}} \|_2^2
    \label{eq:loss}
\end{equation*}
where $P$ denotes output probabilities, $\mathbf{H}^{(l)}$ represents hidden states at layer $l$, and $\beta$ balances the two terms.

\paragraph{Numerical Stability.} Standard \textit{additive} attention masking ($QK^\top + \log M$) introduces gradient singularities ($\nabla M \propto 1/M$) as masks approach zero. To ensure robust training, we implement a custom differentiable FlashAttention kernel. This kernel employs a \textit{multiplicative} formulation that is algebraically equivalent to the standard attention but numerically stable even under BF16 precision (details in Appendix~\ref{app:triton_kernel}).

\section{Experiments}
\label{sec:experiments}

\subsection{Experimental Setup}
\label{sec:exp_setup}

\textbf{Models.}
We evaluate \methodname{} on two widely adopted open-weights models: Llama-3.1-8B-Instruct~\citep{Llama3Herd-2024} and Qwen3-8B~\citep{Qwen3TechnicalReport-2025}.
Both models inherently employ GQA for efficient inference.
Crucially, our method is fully compatible with the GQA structure and achieves significant \textit{further} compression on top of these already efficient architectures, adding only $\sim$0.1\% trainable parameters (see architecture details in Appendix~\ref{app:model_architecture}).

\paragraph{Benchmarks.}
To evaluate the quality of compression, we adopt two standard benchmarks.
LongBench~\citep{LongBenchBilingualMultitask-2024} is used as our primary benchmark for understanding long-contexts, where we select 16 representative subtasks.
See Appendix~\ref{app:longbench_filtering} for detailed configurations and filtering criteria.
We further evaluate on RULER~\citep{RULERWhatsReal-2024}, a synthetic benchmark that systematically probes model performance under compressed long-context inputs across varying sequence lengths and task complexities.

\paragraph{Training Details.}
\methodname{} is trained via self-distillation on our \textbf{Loom} dataset ($\sim$23k samples), strictly decontaminated from evaluation benchmarks (see Appendix~\ref{app:training_data}). The process is cost-effective, taking $<2$ hours on $8\times$A100 GPUs (hyperparameters in Appendix~\ref{app:hyperparameters}).

\paragraph{Baselines.}
Implemented via KVPress~\citep{ExpectedAttentionKV-2025}, we compare \methodname{} against Full Cache and several baselines:
SnapKV~\citep{SnapKVLLMKnows-2024} (uniform budgets);
PyramidKV~\citep{PyramidKVDynamicKV-2025} (layer-wise budget decay);
DuoAttention~\citep{DuoAttentionEfficientLongContext-2024} (offline optimized head patterns);
and AdaKV~\citep{AdaKVOptimizingKV-2025} (specifically \textit{Ada-SnapKV}, in-layer adaptive budgeting).
Additionally, Expected Attention (ExpAttn)~\citep{ExpectedAttentionKV-2025} is included for component analysis.
All experiments follow a question-agnostic~\citep{AdaKVOptimizingKV-2025} setup, where the question is unavailable during context compression.

\subsection{General Long-Context Capabilities}
\label{sec:longbench_results}

\begin{figure}[t]
    \centering
    \includegraphics[width=0.95\linewidth]{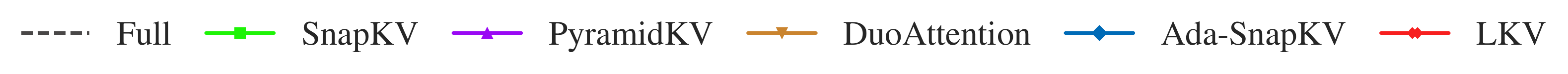}

    \begin{minipage}[b]{0.49\linewidth}
        \centering
        \includegraphics[width=\linewidth]{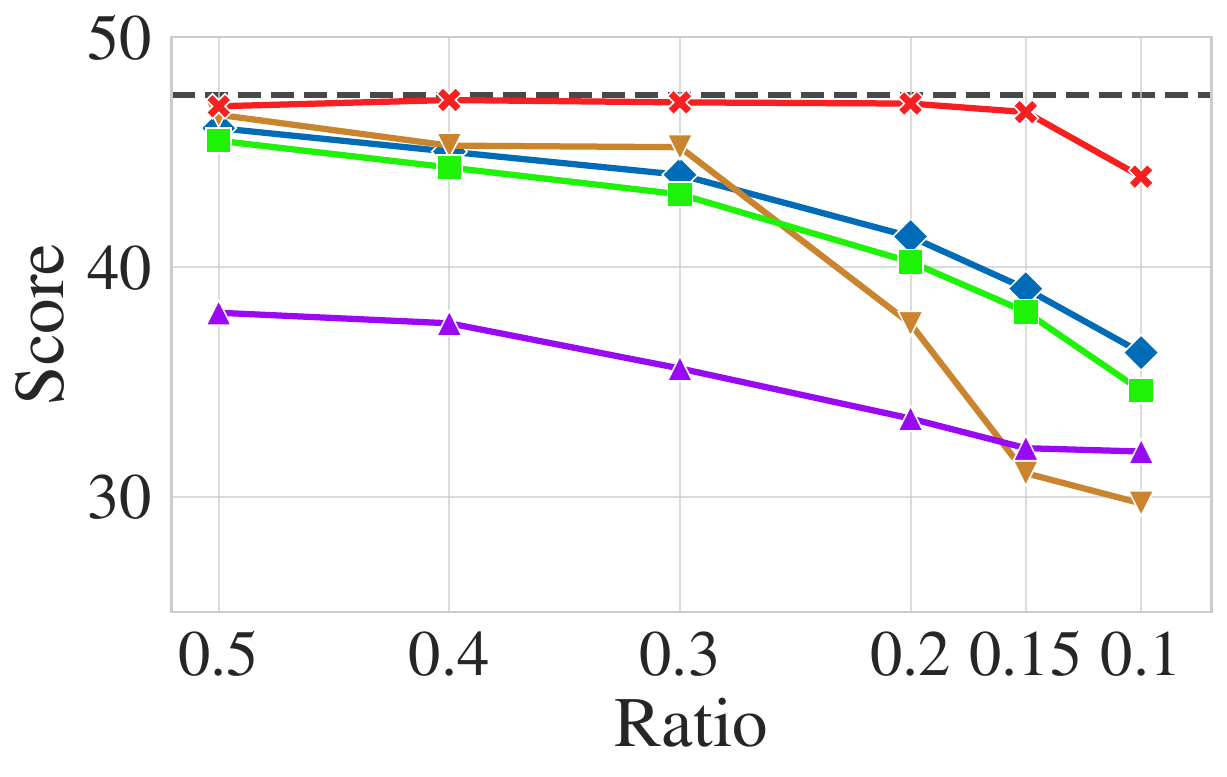}
        \centerline{\scriptsize (a) Llama-3.1-8B-Instruct}
    \end{minipage}
    \hfill
    \begin{minipage}[b]{0.49\linewidth}
        \centering
        \includegraphics[width=\linewidth]{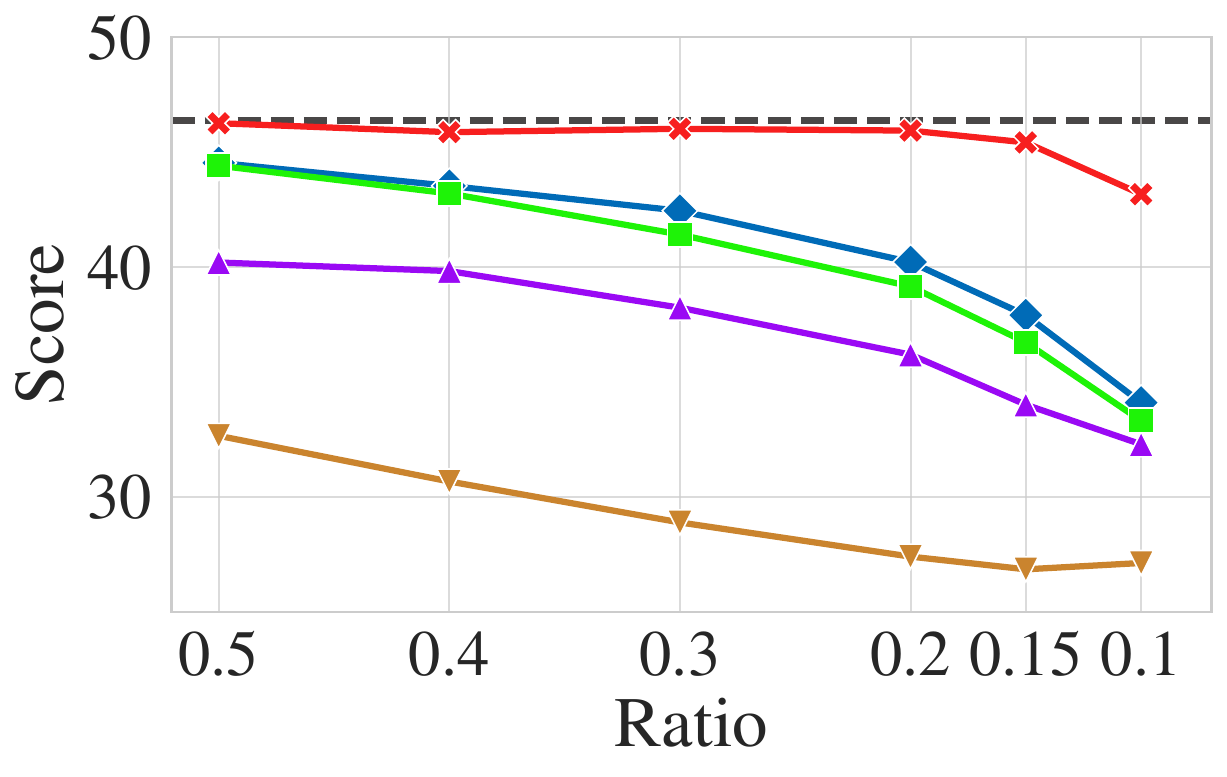}
        \centerline{\scriptsize (b) Qwen3-8B$^{\dagger}$}
    \end{minipage}

    \caption{Performance on LongBench across varying KV cache retention ratios.
        \methodname{} shows robustness in low-resource regimes.}
    \label{fig:performance_curve}
    \raggedright %
    \footnotesize %
    \textit{$^{\dagger}$For Qwen3, we employ an on-the-fly approximation due to the lack of official pre-computed patterns. (see Appendix~\ref{app:baseline_details})}
\end{figure}

We evaluate the comprehensive capabilities of \methodname{} on LongBench.
Table~\ref{tab:longbench} presents the comparison against representative baselines on Llama-3.1-8B-Instruct under a strict 15\% KV budget ($R=0.15$); for the complete task-wise breakdown, please refer to Table~\ref{tab:full_breakdown_llama}.
Figure~\ref{fig:performance_curve} visualizes the performance trends across varying retention ratios.

\paragraph{Overall Performance Comparison.}
\methodname{} achieves an average score of 46.73, consistently outperforming representative baselines.
It recovers 98.4\% of the Full Cache performance (46.73 vs. 47.48), demonstrating virtually lossless context retention.
Compared to the strongest heuristic baseline, Ada-SnapKV, our method leads by a substantial margin of 7.67 points.
This gap highlights the structural advantage of our approach: while heuristic methods relying on local statistics often misallocate resources, \methodname{} learns a global budgeting policy that directly aligns with the generation objective.
We observe consistent superiority on Qwen3-8B as well, with detailed results provided in Table~\ref{tab:full_breakdown_qwen}.

\paragraph{Overcoming Rigid Priors.}
A key advantage of \methodname{} is its ability to break free from hand-crafted structural constraints.
Baselines like PyramidKV and DuoAttention enforce pre-defined patterns—specifically, a rigid layer-wise decay or a fixed offline binary classification.
As shown in Table~\ref{tab:longbench}, these methods suffer severe degradation on retrieval-intensive tasks.
For instance, in the Synthetic category (heavily impacted by the \textit{Passage Retrieval} task, see full breakdown in Table~\ref{tab:full_breakdown_llama}), DuoAttention and PyramidKV drop to 17.75 and 24.99, respectively.
These rigid priors aggressively prune deep layers or specific heads that are critical for retrieving specific information.
In contrast, \methodname{} achieves a high score of 53.38 (matching Full Cache) by automatically identifying and preserving these critical capacities.

\paragraph{Learned vs. Heuristic Adaptation.}
Although Ada-SnapKV introduces adaptivity, its marginal improvement over the uniform SnapKV baseline (39.06 vs. 38.05) exposes the fundamental limitations of heuristic-based allocation.
First, Ada-SnapKV relies on \textit{attention concentration} as a proxy for budget demand.
This heuristic is insufficient because a head's concentration pattern does not necessarily correlate with its task importance, especially when the method is still constrained by a rigid uniform layer-wise budget.
It fails to reallocate resources across layers to preserve deep-context information.
Second, the underlying token selection still depends on accumulated attention scores, which rely on historical statistics rather than predicting future retrieval needs.
\methodname{} overcomes these dual barriers by learning a global policy that directly optimizes the generation objective, unrestricted by layer boundaries or heuristic proxies.

\paragraph{Robustness across Budgets.}
Figure~\ref{fig:performance_curve} illustrates the performance consistency of \methodname{} on both Llama-3.1-8B-Instruct and Qwen3-8B models.
Our method consistently maintains a superior performance-efficiency trade-off compared to baselines.
Notably, the performance advantage remains highly significant as the budget decreases (e.g., at 10\% budget).
This demonstrates that \methodname{} is also exceptionally robust in extreme low-resource regimes.

\begin{table*}[t]
    \centering
    \caption{Main Results on LongBench (Llama-3.1-8B-Instruct).
        Evaluated under 15\% KV budget ($R=0.15$).
    }
    \label{tab:longbench}
    \begin{tabular}{lccccccc}
        \toprule
        Method       & \textbf{Avg.}  & Single-Doc. QA & Multi-Doc. QA  & Summarization  & Few-Shot       & Synthetic      & Code           \\
        \midrule
        Full Cache   & 47.48          & 43.57          & 45.41          & 28.89          & 65.43          & 53.48          & 51.42          \\
        \midrule
        SnapKV       & 38.05          & 27.27          & 34.08          & 22.93          & 58.11          & 40.73          & 50.05          \\
        PyramidKV    & 32.12          & 25.17          & 26.75          & 22.57          & 49.29          & 24.99          & 46.29          \\
        DuoAttention & 31.03          & 21.76          & 25.94          & 21.66          & 50.71          & 17.75          & 50.40          \\
        Ada-SnapKV   & 39.06          & 27.38          & 35.46          & 23.29          & 59.28          & 43.77          & 50.56          \\
        LKV (Ours)   & \textbf{46.73} & \textbf{41.66} & \textbf{42.79} & \textbf{28.76} & \textbf{66.30} & \textbf{53.38} & \textbf{51.25} \\
        \bottomrule
    \end{tabular}
\end{table*}

\subsection{Robustness to Context Length and Compression}
\label{sec:ruler_results}

\begin{figure}[t]
    \centering
    \includegraphics[width=0.95\linewidth]{figures/legend.pdf}
    \begin{minipage}[b]{0.49\linewidth}
        \centering
        \includegraphics[width=\linewidth]{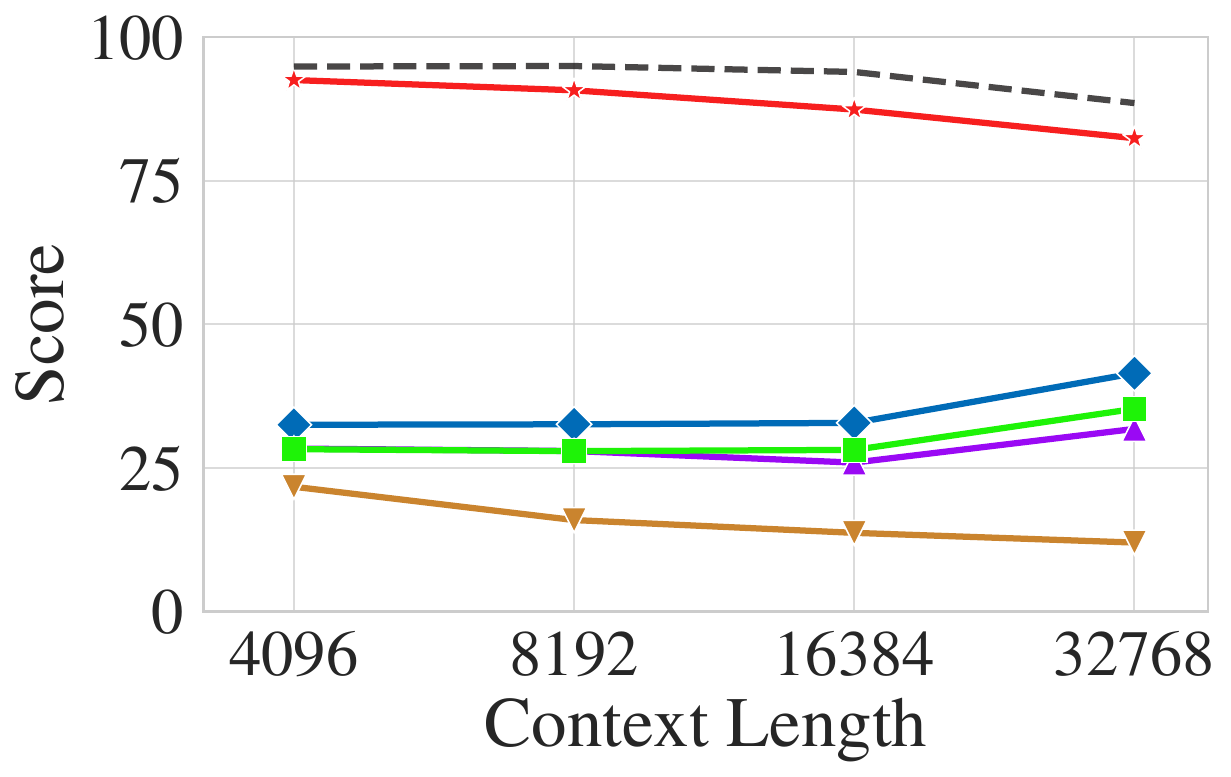}
        \centerline{\footnotesize (a) Varying Length ($R=0.15$)}
    \end{minipage}
    \hfill
    \begin{minipage}[b]{0.49\linewidth}
        \centering
        \includegraphics[width=\linewidth]{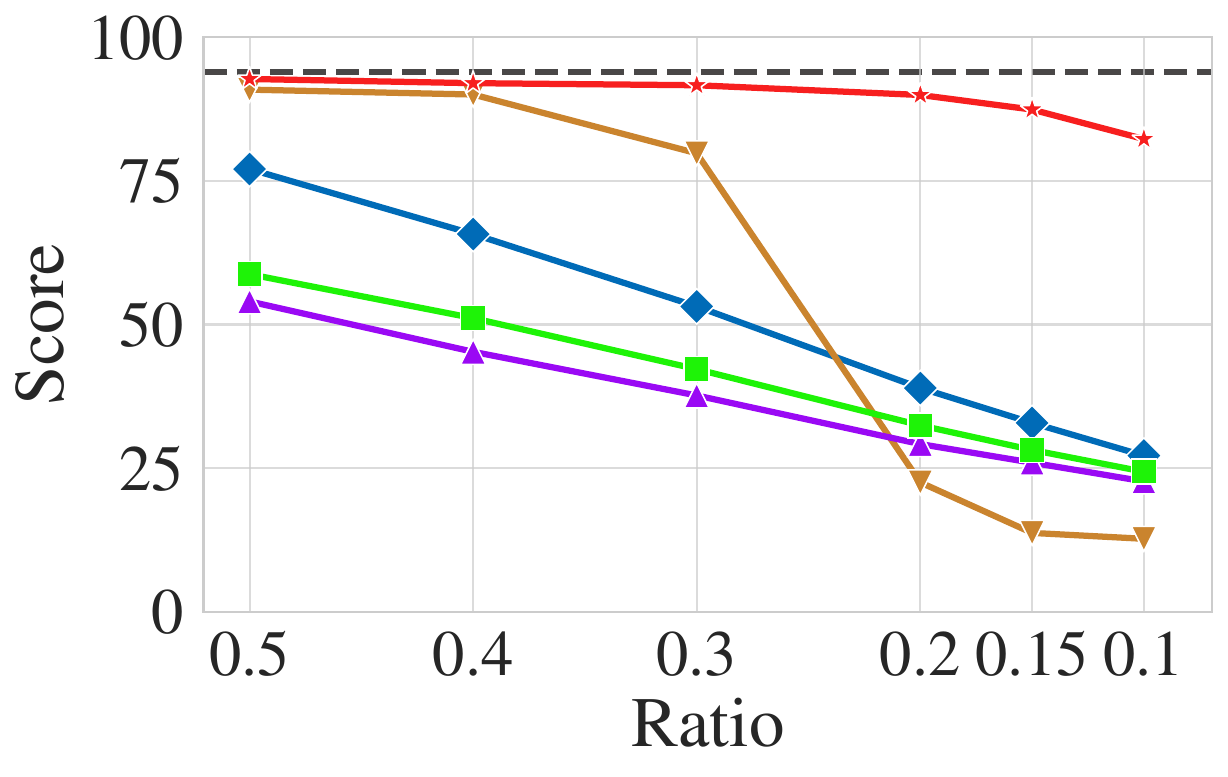}
        \centerline{\footnotesize (b) Varying Ratio ($t=16k$)}
    \end{minipage}
    \caption{Robustness evaluation on RULER (Llama-3.1-8B-Instruct).}
    \label{fig:ruler_analysis}
\end{figure}

We further evaluate the robustness of \methodname{} on Llama-3.1-8B-Instruct by varying the context length $t$ and the global retention ratio $R$.
The results are visualized in Figure~\ref{fig:ruler_analysis}.
Results on Qwen models show consistent trends and are visualized in Figure~\ref{fig:qwen_ruler_results}.

\paragraph{Scalability to Long Contexts.}
Figure~\ref{fig:ruler_analysis}(a) demonstrates performance stability as context extends from 4k to 32k ($R=0.15$).
Despite training within 16k, \methodname{} generalizes robustly to unseen lengths (32k).
We attribute this to the learned policy's transferability: \methodnameH{} captures intrinsic head patterns (e.g., sparse or dense) that remain consistent across lengths, while \methodnameT{} derives token utility strictly from the KV states themselves.
Interestingly, baselines like SnapKV exhibit a slight artifactual rebound at 32k.
This arises because the fixed ratio implies a doubled absolute capacity (e.g., $\sim$4.8k tokens), offering a larger buffer for heuristics to capture scattered information by chance.
Nevertheless, \methodname{} consistently outperforms these methods by a significant margin.

\paragraph{Performance under Extreme Compression.}
Figure~\ref{fig:ruler_analysis}(b) evaluates performance on fixed length inputs ($t=16k$) across varying retention ratios (from 0.5 down to 0.1).
The advantage of \methodname{} becomes more pronounced in low-resource regimes.
Even when the retention ratio is reduced to $R=0.1$, \methodname{} retains a high degree of performance.
While AdaKV performs better than rigid priors (e.g., PyramidKV, DuoAttention) due to its adaptive nature, it still lags behind \methodname{} under extreme compression.
Structural priors like DuoAttention experience significant degradation at lower ratios, failing to adapt their rigid head-pruning strategies to tight budgets.
This confirms that \methodname{} utilizes the limited budget more effectively than both heuristic adaptive methods and static priors.

\subsection{Ablation Study: Dissecting the Gains}
\label{sec:ablation}

To disentangle the contributions of budget allocation (\methodnameH{}) and token selection (\methodnameT{}), we conduct a comprehensive grid analysis on LongBench using Llama-3.1-8B-Instruct (15\% budget).
We combine three budgeting strategies (Uniform, AdaKV, \methodnameH{}) with three token selection policies: SnapKV, \methodnameT{}, and ExpAttn, a recent method that selects tokens by estimating their expected future attention.
Table~\ref{tab:ablation_matrix} reports the macro-average score (Avg.) across all 16 subtasks, along with results on representative retrieval-intensive subsets.

\paragraph{Dominance of Learned Budgeting (\methodnameH{}).}
The results reveal that budget allocation is the primary driver of performance.
While heuristic adaptation (AdaKV) provides only marginal gains over the Uniform baseline, our learned policy (\methodnameH{}) delivers massive improvements across all selectors.
For instance, \methodnameH{} boosts ExpAttn by +8.60 points on average.
Crucially, under uniform budgeting, ExpAttn suffers catastrophic failure on retrieval-heavy tasks (e.g., 17.77 on Synthetic); however, \methodnameH{} acts as a universal enhancer, rescuing it to 45.00.
We observe consistent gains on real-world tasks like Multi-Doc QA (e.g., +7.67 gain for SnapKV), proving that \methodnameH{} effectively corrects the flaws of heuristic selectors.

\paragraph{Superiority of LKV-T Selector.}
Comparing the columns in Table~\ref{tab:ablation_matrix}, our \methodnameT{} selector consistently outperforms existing methods.
Under the Uniform budget, \methodnameT{} achieves 39.74, surpassing both SnapKV (38.05) and ExpAttn (37.12).
When combined with our learned budget, the full \methodname{} system achieves the highest average score of 46.73.
This confirms that \methodnameT{} provides a highly accurate yet efficient (Matrix-free) selection mechanism.

\begin{table}[t]
    \centering
    \caption{Impact of different budgeting and selection strategy combinations.
        Evaluated on LongBench (Llama-3.1-8B-Instruct) under 15\% KV budget.
    }
    \label{tab:ablation_matrix}
    \resizebox{1.0\linewidth}{!}{
        \begin{tabular}{l|l|c|c|c|c}
            \toprule
            Selection & Budgeting                         & \textbf{Avg.}  & $\Delta_{Base}$                            & Multi-Doc QA   & Synthetic      \\
            \midrule
            \multirow{3}{*}{SnapKV}
                      & Uniform                           & 38.05          & -                                          & 34.08          & 40.73          \\
                      & AdaKV                             & 39.06          & \textcolor{blue}{+1.01}                    & 35.46          & 43.77          \\
                      & \cellcolor{gray!10}\methodnameH{} & \textbf{44.06} & \textbf{\textcolor{green!60!black}{+6.01}} & \textbf{41.75} & \textbf{52.87} \\
            \midrule
            \multirow{3}{*}{ExpAttn}
                      & Uniform                           & 37.12          & -                                          & 37.39          & 17.77          \\
                      & AdaKV                             & 38.16          & \textcolor{blue}{+1.04}                    & 37.33          & 23.26          \\
                      & \cellcolor{gray!10}\methodnameH{} & \textbf{45.72} & \textbf{\textcolor{green!60!black}{+8.60}} & \textbf{43.27} & \textbf{45.00} \\
            \midrule
            \multirow{3}{*}{\methodnameT{}}
                      & Uniform                           & 39.74          & -                                          & 40.27          & 22.41          \\
                      & AdaKV                             & 43.45          & \textcolor{blue}{+3.71}                    & 38.83          & 47.76          \\
                      & \cellcolor{gray!10}\methodnameH{} & \textbf{46.73} & \textbf{\textcolor{green!60!black}{+6.99}} & \textbf{42.79} & \textbf{53.38} \\
            \bottomrule
        \end{tabular}
    }
\end{table}

\subsection{Overhead and Memory Analysis}
\label{sec:overhead}

We analyze the computational overhead and memory efficiency of \methodname{} during the prefill phase of Llama-3.1-8B-Instruct ($R=0.15$) on A100-80G.
To accurately measure storage benefits, we employ a custom implementation using \texttt{flash\_attn\_varlen\_func}~\citep{FlashAttentionFastMemoryEfficient-2022}, which flattens the compressed KV cache into a compact format to eliminate padding overhead.

\paragraph{Eviction Overhead and Complexity.}
Table~\ref{tab:prefill_stats} compares the prefill latency at 200k context length (the maximum supported by Full Cache).
DuoAttention achieves the lowest latency due to its static head pruning strategy ($O(1)$ selection complexity).
In comparison, \methodname{} employs a lightweight MLP for dynamic token selection ($O(T)$ complexity).
The measured overhead is marginal: compared to the Full Cache baseline (82.40s), \methodname{} incurs roughly a 7.7\% increase in latency (88.76s).
This is comparable to AdaKV (88.37s), which also performs dynamic filtering.
Considering the significant gains in downstream task performance and memory capacity, this minor construction cost is a favorable trade-off.

\begin{figure}[t]
    \centering
    \includegraphics[width=0.95\linewidth]{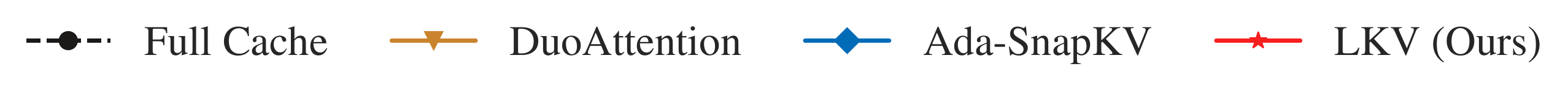}

    \begin{minipage}[b]{0.49\linewidth}
        \centering
        \includegraphics[width=\linewidth]{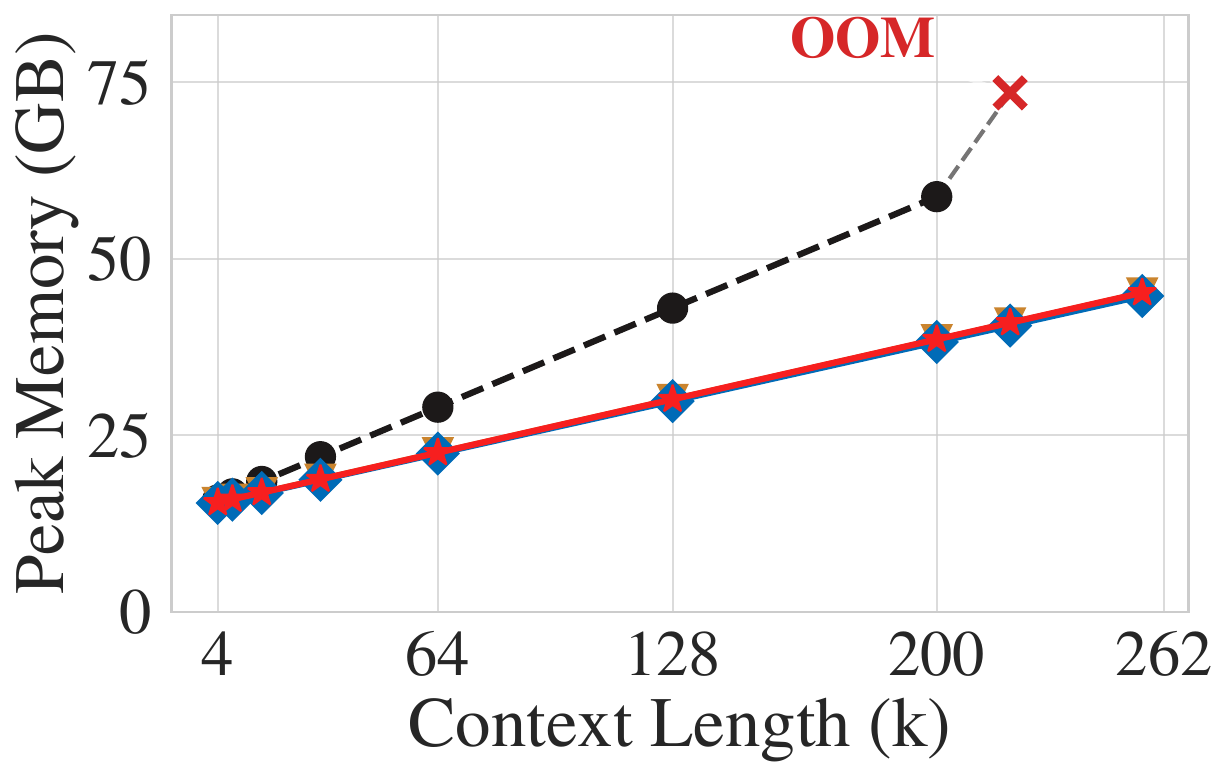}
        \centerline{\footnotesize (a) Peak Memory}
    \end{minipage}
    \hfill
    \begin{minipage}[b]{0.49\linewidth}
        \centering
        \includegraphics[width=\linewidth]{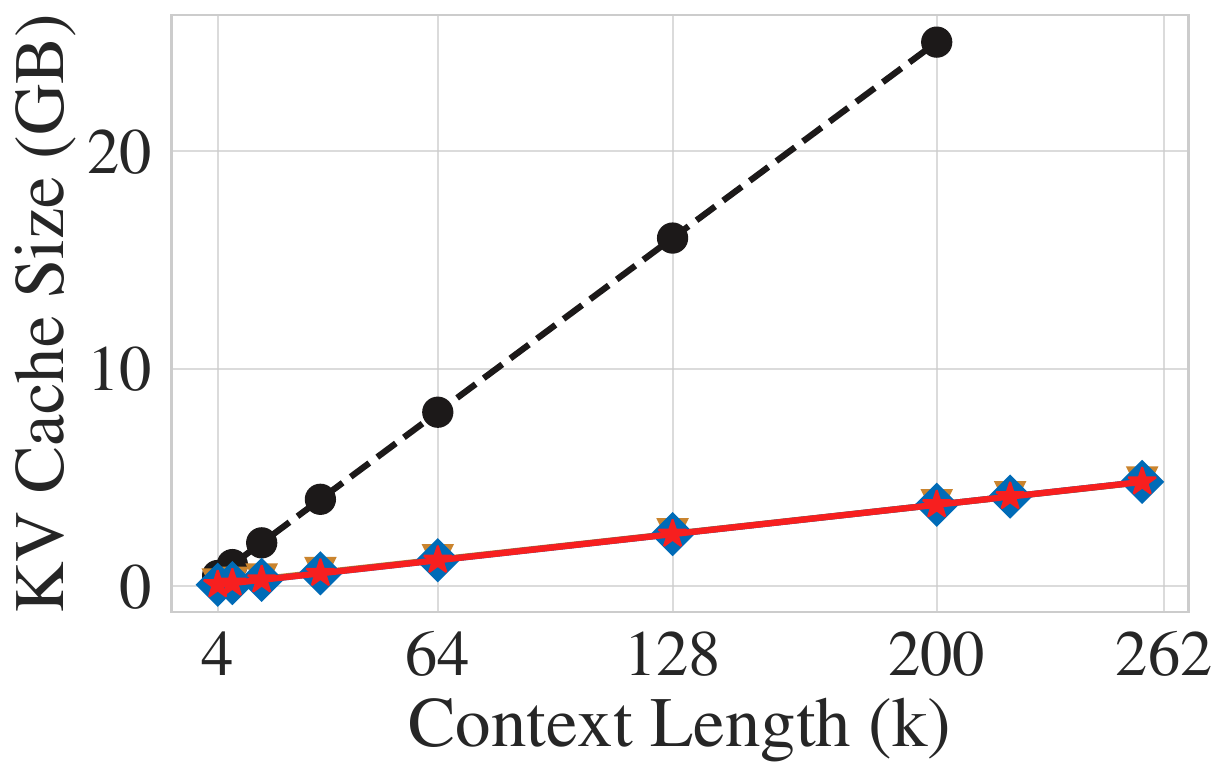}
        \centerline{\footnotesize (b) KV Cache Size}
    \end{minipage}

    \caption{Memory profiling on Llama-3.1-8B-Instruct ($R=0.15$).}
    \label{fig:memory_analysis}
\end{figure}

\paragraph{Memory Efficiency.}
Figure~\ref{fig:memory_analysis} dissects the memory consumption across context lengths.
First, regarding system stability, Full Cache exhibits rapid memory growth, crashing with OOM errors at 225k tokens.
In contrast, \methodname{} successfully scales to 262k and beyond.
This capability stems from the aggressive compression of the KV tensors (Figure~\ref{fig:memory_analysis}b).
By maintaining a 15\% budget, \methodname{} reduces the actual storage cost at 200k length from 25.0 GB (Full) to 3.75 GB (Ours), achieving a 6.6$\times$ reduction.
It is worth noting that the reduction in total peak memory (Figure~\ref{fig:memory_analysis}a) appears less aggressive than the storage compression ratio.
This is because the peak footprint is dominated by static model weights and transient activation buffers, which partially mask the savings from the compressed KV cache.
Nevertheless, the substantial reduction in KV storage is sufficient to extend the maximum context length significantly beyond the hardware limit of Full Cache.

\begin{table}[h]
    \centering
    \caption{Eviction complexity and prefill latency comparison at 200k context length. $W$ denotes the observation window size, and $T$ represents the context length.}
    \label{tab:prefill_stats}
    \resizebox{1.0\linewidth}{!}{
        \begin{tabular}{l|c|c|c}
            \toprule
            Method               & Complexity     & Time (s) & Overhead \\
            \midrule
            Full Cache           & -              & 82.40    & -        \\
            \midrule
            DuoAttention         & $O(1)$         & 87.21    & +5.8\%   \\
            Ada-SnapKV           & $O(W \cdot T)$ & 88.37    & +7.2\%   \\
            \methodname{} (Ours) & $O(T)$         & 88.76    & +7.7\%   \\
            \bottomrule
        \end{tabular}
    }
\end{table}

\section{Conclusion}

\label{sec:conclusion}

In this paper, we reformulate KV cache eviction as an end-to-end differentiable optimization problem, breaking the reliance on rigid heuristics and attention proxies.
By combining global learned budgeting (\methodnameH) and matrix-free token selection (\methodnameT), LKV achieves state-of-the-art fidelity on LongBench and RULER, delivering near-lossless performance with only 15\% retention and a 6.6$\times$ reduction in storage.
Crucially, our analysis identifies global resource allocation as the dominant factor in compression quality, enabling the deployment of extended contexts on commodity hardware with negligible overhead.
While our approach requires a lightweight training phase and specialized kernels (e.g., flattened FlashAttention) to maximize sparse access efficiency, it establishes a robust, data-driven framework for efficient long-context inference.

\section*{Impact Statement}

This paper presents work whose goal is to advance the field of machine learning. There are many potential societal consequences of our work, none of which we feel must be specifically highlighted here.

\bibliography{references}
\bibliographystyle{icml2026}

\newpage
\appendix
\onecolumn

\section{Theoretical Analysis of Soft-TopK Operator}
\label{app:soft_topk}

This appendix establishes the rigorous mathematical foundation for the Soft-TopK operator introduced in Section 3.2. We provide the derivation of the closed-form threshold and analyze its gradient properties to ensure end-to-end differentiability.

\subsection{Derivation of the Closed-Form Solution}
\label{app:soft_topk_derivation}

The derivation follows the smooth approximation framework for Top-K operators explored by~\citet{PostSoftmaxSearchingSmooth-2024}. Our goal is to find a scalar threshold $\lambda(\mathbf{x})$ such that the sum of activated scores equals the target budget $k$:
\begin{equation}
    \sum_{i=1}^n f(x_i - \lambda(\mathbf{x})) = k,
\end{equation}
where the activation function $f(z)$ is based on the Cumulative Distribution Function (CDF) of the Laplace distribution:
\begin{equation}
    f(z) = \frac{1}{2}(1 + \text{sgn}(z)(1 - e^{-|z|})).
\end{equation}
Expanding this function for non-negative and negative inputs, we obtain:
\begin{equation}
    f(z) =
    \begin{cases}
        1 - \frac{1}{2}e^{-z} & \text{if } z \ge 0 \quad \text{(Saturation Region)} \\
        \frac{1}{2}e^{z}      & \text{if } z < 0 \quad \text{(Exponential Region)}
    \end{cases}
\end{equation}
Given input scores $\mathbf{x} \in \mathbb{R}^n$ sorted in descending order ($x_1 \ge \dots \ge x_n$), let $m$ be the split index such that $x_m \ge \lambda(\mathbf{x}) \ge x_{m+1}$. We decompose the summation constraint into two parts: the "head" ($i \le m$) and the "tail" ($i > m$).

For the head elements, where $x_i - \lambda \ge 0$:
\[
    f(x_i - \lambda) = 1 - \frac{1}{2}e^{-(x_i - \lambda)} = 1 - \frac{1}{2}e^{-x_i}e^{\lambda}.
\]
For the tail elements, where $x_i - \lambda < 0$:
\[
    f(x_i - \lambda) = \frac{1}{2}e^{x_i - \lambda} = \frac{1}{2}e^{x_i}e^{-\lambda}.
\]
Substituting these expressions back into the budget constraint yields:
\begin{align}
    k & = \sum_{i=1}^m \left( 1 - \frac{1}{2}e^{-x_i}e^{\lambda} \right) + \sum_{i=m+1}^n \left( \frac{1}{2}e^{x_i}e^{-\lambda} \right)            \\
    k & = m - \frac{1}{2}e^{\lambda} \underbrace{\sum_{i=1}^m e^{-x_i}}_{S_1} + \frac{1}{2}e^{-\lambda} \underbrace{\sum_{i=m+1}^n e^{x_i}}_{S_2},
\end{align}
where $S_1$ and $S_2$ are the partial sums of the exponential scores. To solve for $\lambda$, we introduce the substitution $y = e^{\lambda}$. Multiplying the equation by $2y$ (noting $y > 0$) transforms it into a standard quadratic equation:
\begin{equation}
    S_1 y^2 + 2(k - m)y - S_2 = 0.
\end{equation}
Applying the quadratic formula and selecting the positive root (since $y=e^\lambda$ must be positive, and $S_1, S_2 > 0$):
\begin{equation}
    y = \frac{-2(k-m) + \sqrt{4(k-m)^2 + 4 S_1 S_2}}{2 S_1} = \frac{-(k-m) + \sqrt{(k-m)^2 + S_1 S_2}}{S_1}.
\end{equation}
Finally, taking the natural logarithm gives the closed-form solution:
\begin{equation}
    \lambda(\mathbf{x}) = \ln \left( \frac{\sqrt{(k-m)^2 + S_1 S_2} - (k-m)}{S_1} \right).
    \label{eq:closed_form_lambda_appendix}
\end{equation}

\textbf{Numerical Stability at Boundaries.}
The closed-form solution above relies on the denominator $S_1 = \sum_{i=1}^m e^{-x_i}$ being non-zero. A potential singularity arises only if $m=0$ (selecting zero tokens). Conversely, if $m=n$ (selecting all tokens), $S_2 \to 0$, which is handled by the numerator but suggests a threshold $\to -\infty$.
In our framework, we strictly restrict the target budget $k$ to the open interval $(0, n)$. Specifically, we clamp the retention ratio $R$ such that the derived budget $k$ never strictly reaches $0$ or $n$. This constraint ensures that the split index $m$ satisfies $1 \le m < n$, guaranteeing $S_1 > 0$ and $S_2 > 0$. Consequently, the closed-form solution remains numerically stable and bounded throughout the training process.

\subsection{Gradient Analysis and Differentiability}
\label{app:soft_topk_gradients}

To explicitly demonstrate the end-to-end differentiability of the operator, we derive the gradients of $\lambda$ with respect to the input scores $\mathbf{x}$ and the budget $k$. Instead of differentiating the complex closed-form solution (Eq.~\ref{eq:closed_form_lambda_appendix}) directly, we apply the Implicit Function Theorem to the constraint equation:
\begin{equation}
    F(\lambda, \mathbf{x}, k) = \sum_{i=1}^n f(x_i - \lambda) - k = 0.
\end{equation}
Differentiating implicitly with respect to a specific input score $x_j$, we have:
\[
    \frac{\partial F}{\partial x_j} + \frac{\partial F}{\partial \lambda} \frac{\partial \lambda}{\partial x_j} = 0
    \implies f'(x_j - \lambda) + \left( \sum_{i=1}^n f'(x_i - \lambda) \cdot (-1) \right) \frac{\partial \lambda}{\partial x_j} = 0.
\]
Rearranging terms yields the gradient with respect to the input:
\begin{equation}
    \frac{\partial \lambda}{\partial x_j} = \frac{f'(x_j - \lambda)}{\sum_{i=1}^n f'(x_i - \lambda)}.
\end{equation}
Similarly, differentiating with respect to the budget $k$:
\begin{equation}
    \frac{\partial \lambda}{\partial k} = - \frac{1}{\sum_{i=1}^n f'(x_i - \lambda)}.
\end{equation}
\textbf{Conclusion:} The derivative of the activation function, $f'(z) = \frac{1}{2}e^{-|z|}$, corresponds to the Probability Density Function (PDF) of the Laplace distribution. Since the exponential function is strictly positive for all finite inputs, the denominator $\sum_{i=1}^n f'(x_i - \lambda)$ is always strictly positive (non-zero). This guarantees that the gradients exist and are numerically well-behaved, proving that $\lambda(\mathbf{x})$ is a smooth, differentiable function of both $\mathbf{x}$ and $k$.

\section{Derivation and Analysis of Custom Backward Kernel}
\label{app:triton_kernel}

In Section \ref{subsec:training}, we introduced a custom Triton kernel to address the numerical instability of standard masking mechanisms. This appendix analyzes the gradient stability by comparing the backward passes of the standard log-space formulation and our proposed multiplicative formulation.

\subsection{Instability of Standard Log-Space Masking}
\label{app:triton_kernel_instability}

Standard attention implementations, such as PyTorch's \texttt{F.scaled\_dot\_product\_attention}, typically apply masks as an additive bias in the logit domain.
Let $s_i$ be the raw attention score and $m_i \in (0, 1)$ be the learnable soft mask value for the $i$-th token.
The attention probability scalar $p_i$ is computed as:
\begin{equation}
    p_i = \text{softmax}(s_i + \log m_i) = \frac{e^{s_i + \log m_i}}{\sum_j e^{s_j + \log m_j}}.
\end{equation}
To analyze numerical stability, we examine the gradient flow during backpropagation. In automatic differentiation frameworks, gradients are computed via the Chain Rule through the nodes of the computational graph.
The graph for the mask branch involves a logarithmic operation:
\[
    m_i \xrightarrow{\log} z_i \xrightarrow{+ s_i} a_i \xrightarrow{\text{softmax}} p_i,
\]
where $z_i = \log m_i$ is the bias term.
The gradient of the loss $\mathcal{L}$ with respect to the mask $m_i$ is given by:
\begin{equation}
    \frac{\partial \mathcal{L}}{\partial m_i} = \frac{\partial \mathcal{L}}{\partial z_i} \cdot \frac{\partial z_i}{\partial m_i} = \frac{\partial \mathcal{L}}{\partial z_i} \cdot \frac{1}{m_i}.
\end{equation}
\textbf{The Singularity Problem ($1/m_i$):}
The critical instability arises from the derivative of the logarithm, $\frac{1}{m_i}$.
As the LKV training progresses, the model learns to evict tokens by driving their mask values $m_i \to 0$.
In this regime, the local gradient term $\frac{1}{m_i}$ approaches infinity ($\infty$).
In low-precision floating point formats (e.g., BF16 or FP16), this infinity causes numerical overflow, resulting in \enquote{Inf} or \enquote{NaN} gradients.
This instability occurs at the \enquote{log} node independently of the downstream gradients, causing the optimization to collapse whenever the mask attempts to act as a strict filter (i.e., approaching zero).

\subsection{Proposed Multiplicative Formulation}
\label{app:triton_kernel_equivalence}

To resolve the numerical instability identified in Appendix~\ref{app:triton_kernel_instability}, we propose a multiplicative formulation that operates directly on the probability domain. This method eliminates the problematic logarithm operator while maintaining exact mathematical equivalence to the standard formulation.

\textbf{Formulation.}
The computation proceeds in three steps using the standard FlashAttention tiling pattern:
\begin{enumerate}
    \item \textbf{Raw Softmax:} Compute the standard attention probability on the raw scores $\mathbf{s}$:
          \[
              p_{\text{raw}, i} = \text{softmax}(\mathbf{s})_i = \frac{e^{s_i}}{\sum_j e^{s_j}}.
          \]
    \item \textbf{Multiplicative Masking:} Apply the soft mask $m_i$ element-wise to the raw probabilities:
          \[
              \tilde{p}_i = p_{\text{raw}, i} \cdot m_i.
          \]
    \item \textbf{Re-normalization:} Normalize the masked probabilities to ensure they sum to 1:
          \[
              p_i = \frac{\tilde{p}_i}{\sum_j \tilde{p}_j}.
          \]
\end{enumerate}

\textbf{Proof of Algebraic Equivalence.}
Substituting the expression for $p_{\text{raw}, i}$ into the re-normalization step:
\begin{equation}
    p_i = \frac{\frac{e^{s_i}}{\sum_k e^{s_k}} \cdot m_i}{\sum_j \left( \frac{e^{s_j}}{\sum_k e^{s_k}} \cdot m_j \right)}.
\end{equation}
The common denominator term $\sum_k e^{s_k}$ (the partition function of the raw softmax) appears in both the numerator and the denominator summation, thus cancelling out:
\begin{equation}
    p_i = \frac{e^{s_i} \cdot m_i}{\sum_j e^{s_j} \cdot m_j}.
\end{equation}
Using the identity $x = e^{\log x}$, we can rewrite $m_i$ as $e^{\log m_i}$:
\begin{equation}
    p_i = \frac{e^{s_i} \cdot e^{\log m_i}}{\sum_j e^{s_j} \cdot e^{\log m_j}} = \frac{e^{s_i + \log m_i}}{\sum_j e^{s_j + \log m_j}} = \text{softmax}(s_i + \log m_i).
\end{equation}
\textbf{Conclusion:} Our multiplicative kernel produces output strictly identical to the standard $\text{softmax}(s + \log m)$ formulation during the forward pass, but avoids the explicit computation of $\log m$.

\textbf{Compatibility with FlashAttention.}
Crucially, this formulation is fully compatible with the IO-aware design of FlashAttention~\citep{FlashAttentionFastMemoryEfficient-2022}.
In the standard FlashAttention kernel, the mask is applied as an additive bias ($s_i \leftarrow s_i + \text{bias}_i$) before the exponentiation step in the online softmax algorithm.
In our custom Triton kernel, we simply replace this additive operation with a multiplicative one after exponentiation ($e^{s_i} \leftarrow e^{s_i} \cdot m_i$).
Since this modification only alters the on-chip arithmetic operations (ALU) without affecting the global memory access patterns or the block-wise tiling structure, it preserves the $O(t^2)$ memory savings and IO efficiency of the original FlashAttention.
Consequently, LKV achieves numerical stability with negligible runtime overhead compared to standard kernels.

\subsection{Gradient Stability Analysis}
\label{app:triton_kernel_gradients}

We now analyze the gradient stability of our proposed formulation by examining the backward pass. The fundamental difference lies in how the gradient of the loss $\mathcal{L}$ propagates back to the mask $m_i$.

\textbf{Gradient Flow Comparison.}
To pinpoint the source of stability, we compare the partial derivative $\frac{\partial p_i}{\partial m_i}$ (the sensitivity of the output probability to the mask) in both formulations.

\textbf{1. Standard Log-Space Formulation (Unstable):}
Recall the computational graph: $m_i \xrightarrow{\log} z_i \xrightarrow{+ s_i} \dots \rightarrow p_i$.
Applying the chain rule, the gradient contains an inverse term:
\begin{equation}
    \frac{\partial p_i}{\partial m_i} = \frac{\partial p_i}{\partial z_i} \cdot \frac{\partial z_i}{\partial m_i} = \frac{\partial p_i}{\partial z_i} \cdot \boldsymbol{\frac{1}{m_i}}.
\end{equation}
\textit{Instability:} The term $\frac{1}{m_i}$ acts as a gradient scaling factor. As the model learns to evict tokens ($m_i \to 0$), this factor approaches infinity, causing gradient explosion regardless of the downstream gradient magnitude.

\textbf{2. Proposed Multiplicative Formulation (Stable):}
Recall our new computational graph from Appendix~\ref{app:triton_kernel_equivalence}:
\[
    m_i \xrightarrow{\times p_{\text{raw}, i}} \tilde{p}_i \xrightarrow{\text{Normalize}} p_i.
\]
Here, the mask $m_i$ interacts directly with the bounded raw probability $p_{\text{raw}, i}$ via multiplication. The chain rule yields:
\begin{equation}
    \frac{\partial p_i}{\partial m_i} = \sum_{j} \frac{\partial p_i}{\partial \tilde{p}_j} \cdot \frac{\partial \tilde{p}_j}{\partial m_i}.
\end{equation}
Since the operation is element-wise multiplication ($\tilde{p}_j = p_{\text{raw}, j} \cdot m_j$), the derivative is non-zero only when $j=i$, and crucially:
\begin{equation}
    \frac{\partial \tilde{p}_i}{\partial m_i} = p_{\text{raw}, i}.
\end{equation}
Substituting this back, the gradient flow becomes:
\begin{equation}
    \frac{\partial p_i}{\partial m_i} = \underbrace{\frac{\partial p_i}{\partial \tilde{p}_i}}_{\text{Normalization Jacobian}} \cdot \qquad p_{\text{raw}, i}.
\end{equation}

\textbf{Conclusion on Stability.}
Our formulation effectively replaces the unstable scaling factor $\frac{1}{m_i}$ with the stable factor $p_{\text{raw}, i}$.
\begin{itemize}
    \item \textbf{Boundedness:} The raw probability $p_{\text{raw}, i}$ is the output of a standard softmax, strictly bounded in $(0, 1]$. It represents the intrinsic attention weight of the token before masking.
    \item \textbf{Zero-Preserving:} When a token is evicted ($m_i \to 0$), the gradient does not explode; instead, it is scaled by the token's original importance $p_{\text{raw}, i}$.
\end{itemize}
By shifting the masking operation from the logit domain to the probability domain, we guarantee that the gradient $\nabla_{m}$ remains finite and well-behaved throughout the entire training process, enabling stable optimization even in low-precision (BF16) environments.

\section{Training Implementation Details}
\label{app:training_details}

\subsection{Model Architecture and Parameter Efficiency}
\label{app:model_architecture}

In this section, we detail the architecture and parameter statistics of the LKV modules. The system introduces negligible parameter overhead compared to the backbone LLMs (approximately 0.1\% of the total parameters for an 8B model).

\paragraph{LKV-H (Global Budgeting).}
This module consists of learnable head embeddings and a shared scoring network.
\begin{itemize}
    \item Embeddings: We assign a unique embedding vector $\mathbf{e}^{(l,h)} \in \mathbb{R}^{d_{head}}$ to each KV head ($L \times H_{kv}$ embeddings in total).
    \item Shared Predictor: A single, shared 2-layer MLP processes these embeddings to predict global budget ratios. Since this predictor is shared across all heads, its parameter count is constant.
\end{itemize}

\paragraph{Parameter Analysis.}
We analyze the parameter overhead for both Llama-3.1-8B-Instruct ($L=32, H_{kv}=8$) and Qwen3-8B ($L=36, H_{kv}=8$). The head dimension $d_{head}$ is 128 for both models. Table \ref{tab:mlp_architecture} shows the detailed parameter count for each component.

For Llama-3.1-8B, the LKV-H module comprises $32 \times 8 = 256$ head embeddings plus a single shared MLP, totaling approximately 0.04M parameters. The LKV-T module consists of 256 independent MLPs (one per KV head), each with roughly 33k parameters, summing to 8.45M. The total added overhead is approximately 8.5M parameters, representing 0.11\% of the backbone.

For Qwen3-8B, due to its deeper architecture (36 layers), the overhead is slightly higher but remains negligible. The LKV-H embeddings scale to 288 instances, and the LKV-T module also scales to 288 independent instances. The total added parameters are approximately 9.6M, which corresponds to 0.12\% of the backbone model.

\begin{table}[h]
    \centering
    \caption{Architecture and parameter breakdown of LKV modules. "Config / Arch." details the vector dimension or MLP structure (Input $\to$ Hidden $\to$ Output). All MLPs use SiLU activation. "Unit Params" denotes the parameter count of a single instance.}
    \label{tab:mlp_architecture}
    \begin{small}
        \begin{tabular}{@{}l c c c c c c@{}}
            \toprule
                                 &                     &             & \multicolumn{2}{c}{Llama-3.1-8B} & \multicolumn{2}{c}{Qwen3-8B}                                     \\
            \cmidrule(lr){4-5} \cmidrule(lr){6-7}
            Module               & Config / Arch.      & Unit Params & Count                            & Total Params                 & Count & Total Params              \\
            \midrule
            LKV-H Embeds         & Vector ($d$=128)    & 128         & 256                              & 32,768                       & 288   & 36,864                    \\
            LKV-H Predictor      & $128 \to 64 \to 1$  & 8,320       & 1                                & 8,320                        & 1     & 8,320                     \\
            \midrule
            LKV-T Predictor      & $256 \to 128 \to 1$ & 33,024      & 256                              & 8,454,144                    & 288   & 9,510,912                 \\
            \midrule
            \textbf{Total Added} & -                   & -           & -                                & \textbf{$\approx$ 8.50 M}    & -     & \textbf{$\approx$ 9.56 M} \\
            \bottomrule
        \end{tabular}
    \end{small}
\end{table}

\subsection{Training Data Construction}
\label{app:training_data}

To effectively train the LKV framework for long-context capabilities, we constructed a composite dataset named \textbf{Loom}. The dataset is curated to cover a diverse range of tasks, including multi-hop reasoning, summarization, code completion, and knowledge-based QA.

\textbf{Data Composition.}
The dataset comprises approximately \textbf{23,000 samples} mixed from seven public sources. The composition ratios and task types are detailed in Table~\ref{tab:data_composition}.
\begin{itemize}
    \item \textbf{Long-Context QA \& Reasoning (45\%)}: We integrate \textit{HotpotQA} (30\%) for multi-hop reasoning and \textit{NarrativeQA} (5\%) for long-story comprehension. Additionally, we include \textit{SQuAD} (10\%) formatted as a few-shot task (8-shot) to encourage in-context learning capabilities.
    \item \textbf{Summarization (20\%)}: We utilize \textit{MultiNews} to train the model's ability to aggregate information across long documents.
    \item \textbf{Code Completion (20\%)}: We employ \textit{The Stack} (Python, Java, C++, C\#) to enhance long-range dependency modeling in code, constructing samples by concatenating multiple code blocks.
    \item \textbf{Classification \& Knowledge (15\%)}: We include \textit{TREC} (10\%) for fine-grained classification and \textit{MMLU} (5\%) for general knowledge, both formatted as 5-shot tasks to simulate long-context retrieval scenarios.
\end{itemize}

\textbf{Preprocessing and Filtering.}
All data is formatted into a standardized Chat template. We strictly applied length filtering using the Qwen3-8B tokenizer, retaining only samples with a token length between \textbf{2,000 and 16,000 tokens}. This ensures the model learns effective eviction policies under sufficient KV cache pressure.

\paragraph{Decontamination Strategy.}
We rigorously address the potential risk of data leakage through a multi-level strategy.
First, we strictly excluded \textbf{VCSUM} from our training pipeline, as it is the only dataset in LongBench that utilizes training data for evaluation due to data scarcity.
For the remaining datasets, we ensure strict separation at both the task and sample levels:
\begin{itemize}
    \item \textbf{Limited Task Overlap}: The intersection of tasks is minimal. Our training data covers only a small subset of the tasks found in LongBench. The vast majority of LongBench tasks (e.g., Synthetic tasks) are unseen during training, ensuring that our evaluation reflects true generalization capabilities.
    \item \textbf{Sample-Level Disjointness}: For the few tasks that overlap between our training data and LongBench (e.g., HotpotQA, MultiNews), the specific samples are mutually exclusive. As explicitly stated in the LongBench paper regarding their data extraction process:
          \begin{quote}
              ``Since LLMs may have already been trained on the training set of some of our collected public datasets, to avoid test leakage, we extract data from the \textbf{test sets} of these public datasets, with the exception of VCSUM due to its insufficient data in its test set.''~\citep{LongBenchBilingualMultitask-2024}
          \end{quote}
          Since LKV is trained exclusively on the \textbf{training splits} of these public sources, there is zero overlap in sample instances.
\end{itemize}

\begin{table}[h]
    \centering
    \caption{Composition of the Loom training dataset. All subsets are sourced from their official training splits and filtered to the 2k-16k token range.}
    \label{tab:data_composition}
    \begin{small}
        \begin{tabular}{@{}llcc@{}}
            \toprule
            \textbf{Dataset Source} & \textbf{Task Type}    & \textbf{Ratio} & \textbf{Configuration}        \\
            \midrule
            HotpotQA                & Multi-hop QA          & 30\%           & Fullwiki                      \\
            MultiNews               & Summarization         & 20\%           & -                             \\
            The Stack               & Code Completion       & 20\%           & 3-concat (Py, Java, C\#, C++) \\
            SQuAD                   & Reading Comprehension & 10\%           & 8-shot Few-shot               \\
            TREC                    & Classification        & 10\%           & 5-shot / Fine-grained         \\
            MMLU                    & Knowledge QA          & 5\%            & 5-shot / All subsets          \\
            NarrativeQA             & Long-doc QA           & 5\%            & -                             \\
            \bottomrule
        \end{tabular}
    \end{small}
\end{table}

\subsection{Hyperparameters and Training Schedule}
\label{app:hyperparameters}

We train LKV using the AdamW optimizer with BFloat16 precision to ensure training stability and efficiency. We apply gradient clipping with a maximum norm of 1.0. The training process spans 1,000 steps with a global batch size of 64 (1 sample per GPU $\times$ 8 GPUs $\times$ 8 gradient accumulation steps).

\textbf{Learning Rate Schedule.}
Since the LKV modules (MLPs) are lightweight and initialized from scratch, we utilize a relatively large learning rate of $1.0 \times 10^{-3}$. We apply a polynomial learning rate decay (power=1.0, equivalent to linear decay) down to a final value of $1.0 \times 10^{-4}$, following a 20-step (2\%) warmup phase.

\textbf{Dynamic Annealing Strategy.}
A critical component of our training is the dynamic annealing of the Soft-TopK temperature $\tau$ and the target retention ratio $R$. We implement an exponential decay schedule for both parameters to facilitate a curriculum learning effect—starting with a "soft" selection and high budget, and gradually converging to a "hard" selection with the target sparsity.
For the current step $t$ and total steps $T$, the values are updated as:
\begin{equation}
    \tau_t = \tau_{init} \left( \frac{\tau_{final}}{\tau_{init}} \right)^{t/T}, \quad R_t = R_{init} \left( \frac{R_{final}}{R_{init}} \right)^{t/T}
\end{equation}
Specifically, $\tau$ decays from 1.0 to 0.001, and the training budget ratio $R$ decays from 0.5 (50\% retention) to the target 0.15 (15\% retention). This prevents the model from collapsing into trivial solutions early in training.

\textbf{Loss Configuration.}
We set the weight for the hidden state alignment loss to $\beta=0.5$. The distillation temperature is set to 1.0.

Table~\ref{tab:hyperparameters} summarizes the detailed hyperparameters used in our experiments.

\begin{table}[h]
    \centering
    \caption{Hyperparameters for training LKV on Llama-3.1-8B and Qwen3-8B.}
    \label{tab:hyperparameters}
    \begin{tabular}{@{}ll@{}}
        \toprule
        \textbf{Hyperparameter}      & \textbf{Value}          \\
        \midrule
        \textit{Optimization}        &                         \\
        Optimizer                    & AdamW                   \\
        Precision                    & BFloat16                \\
        Peak Learning Rate           & $1.0 \times 10^{-3}$    \\
        End Learning Rate            & $1.0 \times 10^{-4}$    \\
        LR Scheduler                 & Polynomial (Power=1.0)  \\
        Warmup Steps                 & 20 (2\%)                \\
        Global Batch Size            & 64                      \\
        Gradient Accumulation        & 8 steps                 \\
        Max Gradient Norm            & 1.0                     \\
        Max Steps                    & 1,000                   \\
        \midrule
        \textit{Annealing Schedules} &                         \\
        Temperature Mode             & Exponential Decay       \\
        Temperature Range ($\tau$)   & $1.0 \rightarrow 0.001$ \\
        Budget Ratio Mode            & Exponential Decay       \\
        Budget Ratio Range ($R$)     & $0.5 \rightarrow 0.15$  \\
        \midrule
        \textit{Loss Coefficients}   &                         \\
        Hidden Loss Weight ($\beta$) & 0.5                     \\
        Distillation Temp            & 1.0                     \\
        \bottomrule
    \end{tabular}
\end{table}

\section{Evaluation Setup Details}
\label{app:evaluation_setup}

\subsection{Benchmark Filtering and Configuration}
\label{app:longbench_filtering}

To ensure a fair and unified comparison with prior state-of-the-art methods, our evaluation on LongBench strictly follows the settings adopted by SnapKV~\citep{SnapKVLLMKnows-2024} and Ada-KV~\citep{AdaKVOptimizingKV-2025}.

\textbf{Task Selection.}
LongBench consists of 21 subtasks covering multi-language scenarios. Following standard practice in KV cache compression literature, we select the 16 English and Code tasks for evaluation. The 5 excluded tasks are exclusively Chinese-language datasets: \textit{MultiFieldQA-zh}, \textit{DuReader}, \textit{VCSUM}, \textit{LSHT}, and \textit{PassageRetrieval-zh}.
We exclude these tasks for two reasons:
\begin{enumerate}
    \item \textbf{Baseline Alignment:} Representative baselines (e.g., SnapKV, PyramidKV, AdaKV) report results on this specific subset of 16 tasks. Adhering to this protocol guarantees an apples-to-apples comparison.
    \item \textbf{Modality Focus:} The base models (Llama-3.1-8B-Instruct and Qwen3-8B) are trained primarily on English. Evaluating on English tasks ensures a fair comparison of model capabilities in their primary modality.
\end{enumerate}

\textbf{Included Tasks.}
Table~\ref{tab:longbench_tasks} lists the details of the 16 evaluated tasks, categorizing them into Single-Document QA, Multi-Document QA, Summarization, Few-Shot Learning, Synthetic, and Code tasks. The metric for code generation is Edit Similarity (Edit Sim), while others use F1 Score, ROUGE-L, or Accuracy.

\begin{table}[h]
    \centering
    \caption{Details of the 16 LongBench tasks used in our evaluation. This selection aligns with the evaluation protocols of SnapKV and AdaKV.}
    \label{tab:longbench_tasks}
    \begin{small}
        \begin{tabular}{@{}lllc@{}}
            \toprule
            \textbf{Category}                  & \textbf{Task}       & \textbf{Metric} & \textbf{Language} \\
            \midrule
            \multirow{3}{*}{Single-Doc QA}     & NarrativeQA         & F1              & EN                \\
                                               & Qasper              & F1              & EN                \\
                                               & MultiFieldQA-en     & F1              & EN                \\
            \midrule
            \multirow{3}{*}{Multi-Doc QA}      & HotpotQA            & F1              & EN                \\
                                               & 2WikiMQA            & F1              & EN                \\
                                               & MuSiQue             & F1              & EN                \\
            \midrule
            \multirow{3}{*}{Summarization}     & GovReport           & ROUGE-L         & EN                \\
                                               & QMSum               & ROUGE-L         & EN                \\
                                               & MultiNews           & ROUGE-L         & EN                \\
            \midrule
            \multirow{3}{*}{Few-Shot Learning} & TREC                & Accuracy        & EN                \\
                                               & TriviaQA            & F1              & EN                \\
                                               & SAMSum              & ROUGE-L         & EN                \\
            \midrule
            \multirow{2}{*}{Synthetic}         & PassageCount        & Accuracy        & EN                \\
                                               & PassageRetrieval-en & Accuracy        & EN                \\
            \midrule
            \multirow{2}{*}{Code}              & LCC                 & Edit Sim        & Python/C\#/Java   \\
                                               & RepoBench-P         & Edit Sim        & Python/Java       \\
            \bottomrule
        \end{tabular}
    \end{small}
\end{table}

\subsection{Baseline Implementation Details}
\label{app:baseline_details}

We implement all baselines using KVPress~\citep{ExpectedAttentionKV-2025} for unified evaluation.

\paragraph{DuoAttention Implementation.}
DuoAttention~\citep{DuoAttentionEfficientLongContext-2024} prunes \enquote{Streaming Heads} while retaining \enquote{Retrieval Heads}:
\begin{itemize}
    \item \textbf{Llama-3.1-8B}: We utilize the official pre-computed attention patterns provided by the authors to identify head types.
    \item \textbf{Qwen3-8B (On-the-Fly Approximation)}: Since Qwen3-8B was released after DuoAttention and lacks official compatible patterns, we adopt an \textit{on-the-fly} estimation method proposed by~\citet{ExpectedAttentionKV-2025}. Instead of expensive offline calibration, we synthesize attention patterns by probing the model with random context samples. By measuring the attention distribution on these synthetic inputs, we dynamically classify heads as retrieval or streaming, effectively extending DuoAttention's structural prior to Qwen3 without compatibility issues.
\end{itemize}

\paragraph{Other Baselines.}
We adhere to standard configurations: SnapKV uses the recommended observation window size; PyramidKV follows the standard layer-wise decay profile; and Ada-SnapKV employs attention concentration scores for adaptive budgeting.

\section{Additional Experimental Results}
\label{app:additional_results}

\subsection{Detailed LongBench Breakdown}
\label{app:longbench_breakdown}

In this section, we provide the fine-grained performance breakdown for all 16 evaluated LongBench subtasks.
Tables~\ref{tab:full_breakdown_llama} and \ref{tab:full_breakdown_qwen} present the results for Llama-3.1-8B-Instruct and Qwen3-8B, respectively, under a strict 15\% KV budget ($R=0.15$).

\textbf{Analysis of Structural Priors.}
The breakdown reveals why heuristic methods with rigid structural priors suffer in specific domains.
For instance, on the \textit{PassageRetrieval-en} (PRe) task using Llama-3.1-8B-Instruct (Table~\ref{tab:full_breakdown_llama}), DuoAttention and PyramidKV drop to scores of 29.00 and 44.65, respectively, compared to the Full Cache score of 99.50. This confirms that static priors (e.g., heavy penalization of distant tokens or specific heads) aggressively evict needle-in-a-haystack information.
In contrast, LKV achieves a score of 97.00, demonstrating its ability to dynamically identify and retain critical retrieval tokens regardless of their position.

\begin{table*}[h!]
    \centering
    \caption{Llama-3.1-8B-Instruct: Detailed performance breakdown on 16 LongBench tasks ($R=0.15$).}
    \label{tab:full_breakdown_llama}
    \resizebox{\textwidth}{!}{
        \begin{tabular}{lccccccccccccccccc}
            \toprule
            Method       & \textbf{Avg.}  & \rotatebox{45}{NrtvQA} & \rotatebox{45}{Qasper} & \rotatebox{45}{MF-en} & \rotatebox{45}{HotpotQA} & \rotatebox{45}{2WikiMQA} & \rotatebox{45}{Musique} & \rotatebox{45}{GovReport} & \rotatebox{45}{QMSum} & \rotatebox{45}{MultiNews} & \rotatebox{45}{TREC} & \rotatebox{45}{TriviaQA} & \rotatebox{45}{SAMSum} & \rotatebox{45}{PCount} & \rotatebox{45}{PRe} & \rotatebox{45}{Lcc} & \rotatebox{45}{RB-P} \\
            \midrule
            Full Cache   & 47.48          & 32.38                  & 45.22                  & 53.12                 & 56.84                    & 48.04                    & 31.35                   & 34.63                     & 25.25                 & 26.78                     & 68.50                & 90.79                    & 36.99                  & 7.45                   & 99.50               & 50.88               & 51.96                \\
            SnapKV       & 38.05          & 26.73                  & 24.81                  & 30.28                 & 48.35                    & 29.48                    & 24.40                   & 26.02                     & 20.82                 & 21.94                     & 50.00                & 91.81                    & 32.53                  & 6.46                   & 75.00               & 49.95               & 50.16                \\
            PyramidKV    & 32.12          & 22.12                  & 23.97                  & 29.43                 & 39.05                    & 27.38                    & 13.82                   & 25.44                     & 20.41                 & 21.86                     & 38.50                & 87.96                    & 21.40                  & 5.33                   & 44.65               & 50.11               & 42.47                \\
            DuoAttention & 31.03          & 19.63                  & 16.41                  & 29.23                 & 38.09                    & 22.68                    & 17.04                   & 24.10                     & 17.94                 & 22.93                     & 39.50                & 89.12                    & 23.51                  & 6.50                   & 29.00               & 49.34               & 51.46                \\
            Ada-SnapKV   & 39.06          & 25.98                  & 24.07                  & 32.10                 & 49.85                    & 32.58                    & 23.94                   & 25.93                     & 21.89                 & 22.05                     & 52.00                & \textbf{92.16}           & 33.68                  & 7.54                   & 80.00               & \textbf{50.18}      & 50.94                \\
            LKV (Ours)   & \textbf{46.73} & \textbf{31.87}         & \textbf{43.16}         & \textbf{49.94}        & \textbf{57.16}           & \textbf{42.18}           & \textbf{29.04}          & \textbf{34.44}            & \textbf{25.04}        & \textbf{26.79}            & \textbf{69.50}       & 91.80                    & \textbf{37.60}         & \textbf{9.75}          & \textbf{97.00}      & 49.99               & \textbf{52.50}       \\
            \bottomrule
        \end{tabular}
    }
\end{table*}

\begin{table*}[h!]
    \centering
    \caption{Qwen3-8B: Detailed performance breakdown on 16 LongBench tasks ($R=0.15$). Note that DuoAttention utilizes on-the-fly approximation.}
    \label{tab:full_breakdown_qwen}
    \resizebox{\textwidth}{!}{
        \begin{tabular}{lccccccccccccccccc}
            \toprule
            Method       & \textbf{Avg.}  & \rotatebox{45}{NrtvQA} & \rotatebox{45}{Qasper} & \rotatebox{45}{MF-en} & \rotatebox{45}{HotpotQA} & \rotatebox{45}{2WikiMQA} & \rotatebox{45}{Musique} & \rotatebox{45}{GovReport} & \rotatebox{45}{QMSum} & \rotatebox{45}{MultiNews} & \rotatebox{45}{TREC} & \rotatebox{45}{TriviaQA} & \rotatebox{45}{SAMSum} & \rotatebox{45}{PCount} & \rotatebox{45}{PRe} & \rotatebox{45}{Lcc} & \rotatebox{45}{RB-P} \\
            \midrule
            Full Cache   & 46.37          & 28.99                  & 40.06                  & 51.98                 & 56.93                    & 42.99                    & 32.29                   & 33.57                     & 23.98                 & 24.72                     & 71.50                & 88.69                    & 39.74                  & 2.75                   & 100.00              & 50.71               & 53.03                \\
            SnapKV       & 36.72          & 21.59                  & 22.43                  & 27.02                 & 39.62                    & 29.90                    & 21.17                   & 27.42                     & 19.76                 & 19.23                     & 49.50                & \textbf{89.44}           & 38.37                  & 3.26                   & 76.00               & 48.52               & 54.26                \\
            PyramidKV    & 34.01          & 18.42                  & 22.38                  & 25.25                 & 31.97                    & 28.17                    & 17.31                   & 26.33                     & 19.29                 & 19.34                     & 43.50                & 87.51                    & 37.16                  & \textbf{4.57}          & 60.50               & 48.78               & 53.68                \\
            DuoAttention & 26.85          & 11.21                  & 13.07                  & 24.06                 & 28.48                    & 25.27                    & 9.26                    & 22.37                     & 17.05                 & 20.63                     & 30.50                & 84.74                    & 35.65                  & 0.42                   & 7.50                & 45.81               & 53.59                \\
            Ada-SnapKV   & 37.90          & 21.40                  & 23.12                  & 27.68                 & 42.16                    & 32.09                    & 20.83                   & 27.02                     & 20.13                 & 19.73                     & 54.00                & 89.11                    & \textbf{39.30}         & 3.97                   & 83.50               & \textbf{48.88}      & 53.51                \\
            LKV (Ours)   & \textbf{45.41} & \textbf{24.06}         & \textbf{38.23}         & \textbf{50.29}        & \textbf{57.60}           & \textbf{43.59}           & \textbf{31.37}          & \textbf{32.60}            & \textbf{24.10}        & \textbf{24.26}            & \textbf{69.50}       & 88.77                    & 38.54                  & 3.81                   & \textbf{100.00}     & 45.04               & \textbf{54.88}       \\
            \bottomrule
        \end{tabular}
    }
\end{table*}

\begin{table}[h]
    \centering
    \caption{Llama-3.1-8B-Instruct: Average LongBench scores across varying retention ratios ($R$).}
    \label{tab:llama_varying_ratio}
    \begin{tabular}{lcccccc}
        \toprule
        \textbf{Method} & \textbf{R=0.5} & \textbf{R=0.4} & \textbf{R=0.3} & \textbf{R=0.2} & \textbf{R=0.15} & \textbf{R=0.1} \\
        \midrule
        SnapKV          & 45.49          & 44.33          & 43.16          & 40.22          & 38.05           & 34.62          \\
        PyramidKV       & 38.01          & 37.55          & 35.59          & 33.41          & 32.12           & 31.97          \\
        DuoAttention    & 46.62          & 45.28          & 45.21          & 37.53          & 31.03           & 29.72          \\
        Ada-SnapKV      & 46.04          & 45.02          & 44.01          & 41.32          & 39.06           & 36.28          \\
        LKV (Ours)      & \textbf{46.99} & \textbf{47.26} & \textbf{47.16} & \textbf{47.11} & \textbf{46.73}  & \textbf{43.93} \\
        \bottomrule
    \end{tabular}
\end{table}

\begin{table}[h]
    \centering
    \caption{Qwen3-8B: Average LongBench scores across varying retention ratios ($R$).}
    \label{tab:qwen_varying_ratio}
    \begin{tabular}{lcccccc}
        \toprule
        \textbf{Method} & \textbf{R=0.5} & \textbf{R=0.4} & \textbf{R=0.3} & \textbf{R=0.2} & \textbf{R=0.15} & \textbf{R=0.1} \\
        \midrule
        SnapKV          & 44.41          & 43.20          & 41.41          & 39.16          & 36.72           & 33.31          \\
        PyramidKV       & 40.20          & 39.82          & 38.23          & 36.19          & 34.01           & 32.29          \\
        DuoAttention    & 32.66          & 30.67          & 28.89          & 27.40          & 26.85           & 27.13          \\
        Ada-SnapKV      & 44.53          & 43.53          & 42.44          & 40.21          & 37.90           & 34.10          \\
        LKV (Ours)      & \textbf{46.25} & \textbf{45.86} & \textbf{46.01} & \textbf{45.93} & \textbf{45.41}  & \textbf{43.17} \\
        \bottomrule
    \end{tabular}
\end{table}

\subsection{Qwen3-8B performance on RULER benchmark}
\label{app:qwen_ruler}

\begin{figure}[h]
    \centering
    \includegraphics[width=0.7\linewidth]{figures/legend.pdf}

    \begin{minipage}[b]{0.48\linewidth}
        \centering
        \includegraphics[width=0.7\linewidth]{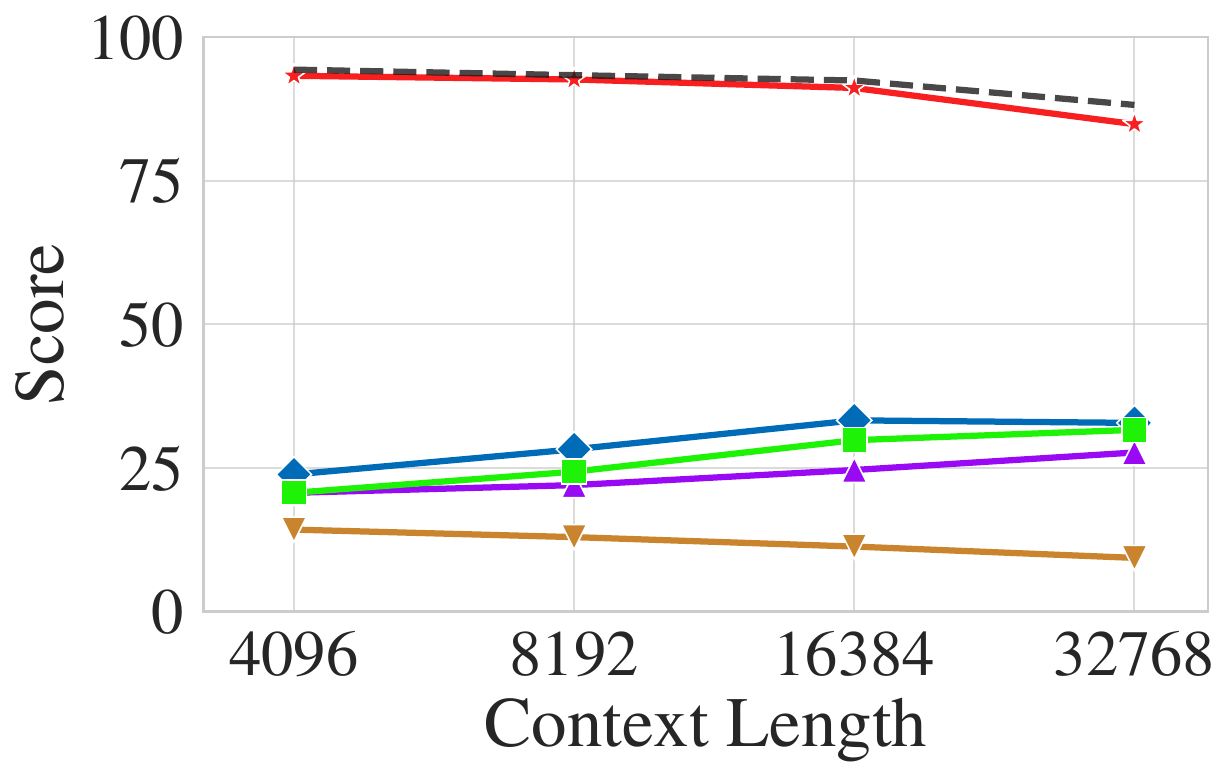}
        \centerline{\scriptsize (a) Varying Length ($R=0.15$)}
    \end{minipage}
    \hfill
    \begin{minipage}[b]{0.48\linewidth}
        \centering
        \includegraphics[width=0.7\linewidth]{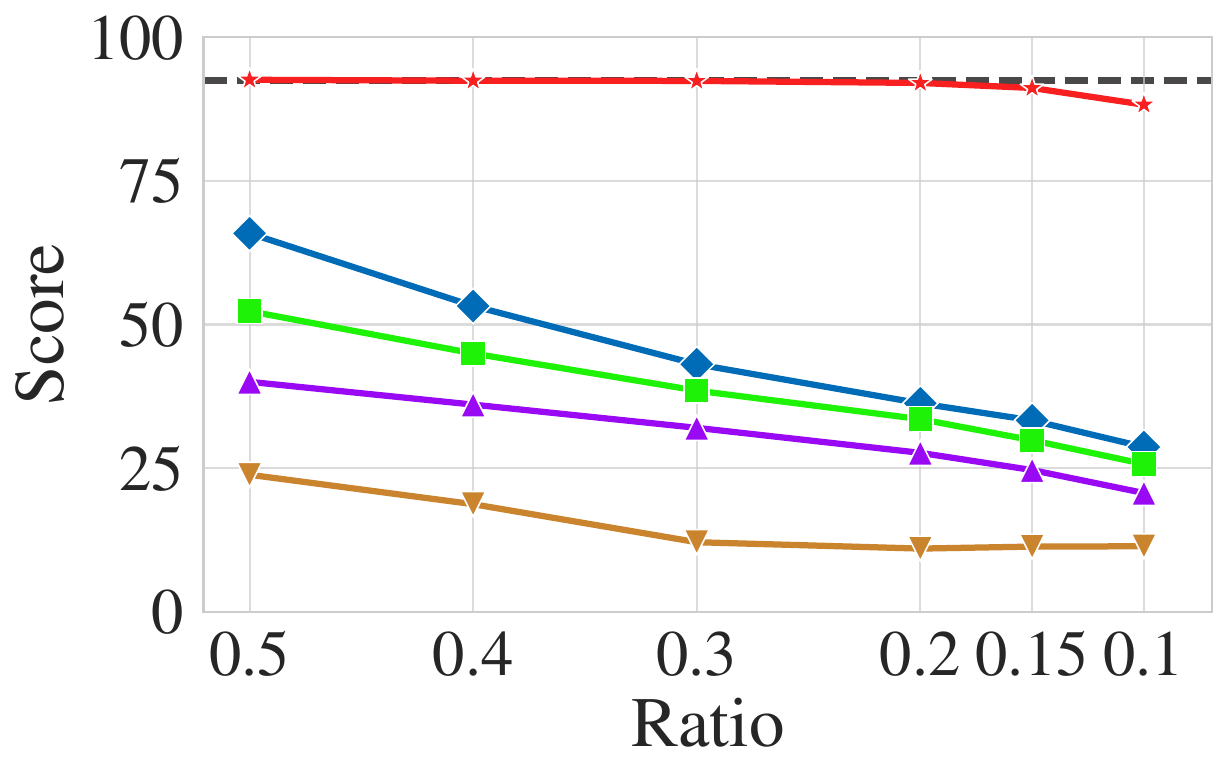}
        \centerline{\scriptsize (b) Varying Ratio ($t=16k$)}
    \end{minipage}

    \caption{
        Qwen3-8B performance on RULER.
        \textbf{(a)} Effective context length evaluation up to 32k tokens.
        \textbf{(b)} Performance degradation analysis under aggressive compression ratios.
        \methodname{} maintains superior stability across both dimensions.
    }
    \label{fig:qwen_ruler_results}
\end{figure}

To verify the generalization of LKV across different architectures, we further evaluate Qwen3-8B on the RULER benchmark.
Figure~\ref{fig:qwen_ruler_results} visualizes the results under two critical settings:
(a) \textbf{Length Scalability}: Performance across increasing context lengths (4k, 8k, 16k, 32k) with a fixed retention ratio ($R=0.15$).
(b) \textbf{Compression Robustness}: Performance across varying retention ratios ($R \in [0.1, 0.5]$) at a fixed context length of 16k.

Consistent with the findings on Llama-3.1-8B-Instruct, LKV demonstrates superior robustness compared to baselines. Notably, at 32k context length, heuristic methods like SnapKV experience significant degradation, whereas LKV maintains high effective context retention.

\end{document}